\documentclass{article}

\PassOptionsToPackage{round}{natbib}

\usepackage[final]{neurips_2024}

\usepackage[utf8]{inputenc}
\usepackage[T1]{fontenc}
\usepackage[dvipsnames]{xcolor}
\usepackage{hyperref}
\hypersetup{
             colorlinks=true,
             linkcolor=Thistle,
             citecolor=Thistle,
             urlcolor=Thistle}
\usepackage{url}            
\usepackage{booktabs}       
\usepackage{amsfonts}       
\usepackage{nicefrac}       
\usepackage{microtype}      
\usepackage{xcolor}     

\usepackage{enumitem}
\usepackage{wrapfig}
\usepackage{blindtext}
\usepackage{amsmath}
\usepackage{amsfonts}
\usepackage{amssymb}
\usepackage{amsthm}
\usepackage{bm}
\usepackage{mathtools}
\usepackage{comment}
\usepackage{graphicx}
\usepackage[capitalize,nameinlink]{cleveref}
\Crefname{section}{\S\hspace{-1mm}}{\S\hspace{-0.5mm}}
\Crefname{appendix}{App.}{Apps.}
\usepackage{mdframed}
\definecolor{theoremcolor}{rgb}{0.94, 0.94, 0.94}
\definecolor{examplecolor}{rgb}{1, 1, 1.0}
\mdfsetup{
    backgroundcolor=theoremcolor,
    linewidth=0pt,
}
\usepackage{xfrac}
\newmdtheoremenv[linewidth=0pt,innerleftmargin=4pt,innerrightmargin=4pt]{definition}{Definition}
\newmdtheoremenv[linewidth=0pt,innerleftmargin=4pt,innerrightmargin=4pt]{proposition}{Proposition}
\newmdtheoremenv[linewidth=0pt,innerleftmargin=0pt,innerrightmargin=0pt,backgroundcolor=examplecolor]{example}{Example}
\newmdtheoremenv{corollary}{Corollary}
\newmdtheoremenv{theorem}{Theorem}
\newmdtheoremenv{lemma}{Lemma}

\title{Reranking Laws for Language Generation: \\ A Communication-Theoretic Perspective}

\author{%
    \textbf{António Farinhas}$^{1,2}$ \quad
    \textbf{Haau-Sing Li}$^{2,3}$ \quad
    \textbf{André F. T. Martins}$^{1,2,4,5}$
    \\
    $^1$Instituto Superior Técnico, Universidade de Lisboa \quad
    $^2$Instituto de Telecomunicações\\
    $^3$Ubiquitous Knowledge Processing Lab, TU Darmstadt \quad
    $^4$ELLIS Unit Lisbon \quad
    $^5$Unbabel\\
    {\small \texttt{\{antonio.farinhas,andre.t.martins\}@tecnico.ulisboa.pt}, \ \texttt{hli@ukp.tu-darmstadt.de}}
}

\begin{document}

\maketitle

\begin{abstract}
    To ensure large language models (LLMs) are used safely, one must reduce their propensity to hallucinate or to generate unacceptable answers. 
    A simple and often used strategy is to first let the LLM generate multiple hypotheses and then employ a reranker to choose the best one. 
    In this paper, we draw a parallel between this strategy and the  use of  redundancy to decrease the error rate in noisy communication channels.
    We conceptualize the generator as a sender transmitting multiple descriptions of a message through parallel noisy channels.
    The receiver decodes the message by ranking the (potentially corrupted) descriptions and selecting the one found to be most reliable. 
    We provide conditions under which this  protocol is asymptotically error-free (\textit{i.e.},  yields an acceptable answer almost surely) even in scenarios where the reranker is imperfect (governed by Mallows or Zipf-Mandelbrot models) and the channel distributions are statistically dependent.  
    We use our framework to obtain reranking laws which we validate empirically on two real-world tasks using LLMs: text-to-code generation with DeepSeek-Coder 7B and machine translation of medical data with TowerInstruct 13B.
\end{abstract}

\section{Introduction}
\label{sec:Introduction}

Large language models (LLMs) have shown remarkable performance across many tasks in natural language processing, computer vision, and speech recognition. 
Despite their capabilities, instances of hallucinations and other critical errors occasionally arise, casting doubt on the reliability of their predictions, without clear indication of when and how badly they might fail \citep{ji2023survey, guerreiro2023hallucinations}.
This is particularly concerning as these models are increasingly used in high-stakes applications such as those within the medical or legal domains \citep{hung-etal-2023-walking} or as agents that can perform multiple tasks, including generating and executing code \citep{wang2024executable}.

The most common mitigation strategy is to ``steer'' the LLM with the aid of a reward model or directly from human preferences, either at training time \citep{stiennon2020learning,yuan2024rrhf,rafailov2024direct} or during  decoding  \citep{liu2024decoding,huang2024deal}. 
A simple and effective decoding-time strategy is first to generate multiple hypotheses and then use a reranker to select the most appropriate one. 
Several generation techniques used with modern LLMs, including {voting procedures} \citep{borgeaud-emerson-2020-leveraging, wang2023selfconsistency, lievin-2024-can, shi-etal-2022-natural}, minimum Bayes risk decoders \citep{eikema-aziz-2020-map,freitag2022high}, {quality-aware decoders} \citep{fernandes-etal-2022-quality}, or other types of {hypothesis ensembling/reranking} techniques \citep{farinhas-etal-2023-empirical, ni2023lever, bertsch-etal-2023-mbr, li2024docefindingsweetspot}, embody this idea.
An essential aspect of these procedures is that they all add \textbf{redundancy} as an intermediate step (by generating multiple hypotheses) to increase the chances of returning an acceptable answer as the final output. 

The idea of adding redundancy to decrease the error rate in noisy channels is a cornerstone of \textbf{communication theory}, more specifically in forward error correction methods. 
In its simplest form---repetition codes---a message block is sent multiple times, and the decoder uses some form of majority voting to recover the original message with high probability \citep{mackay2002information, cover2012elements}.
The same idea underlies more sophisticated error-correcting codes \citep{hamming1950error,reed1960polynomial,gallager1962low,berrou1993near}.

\begin{figure}[t]
\begin{center}
\centerline{
\includegraphics[width=1.0\columnwidth]{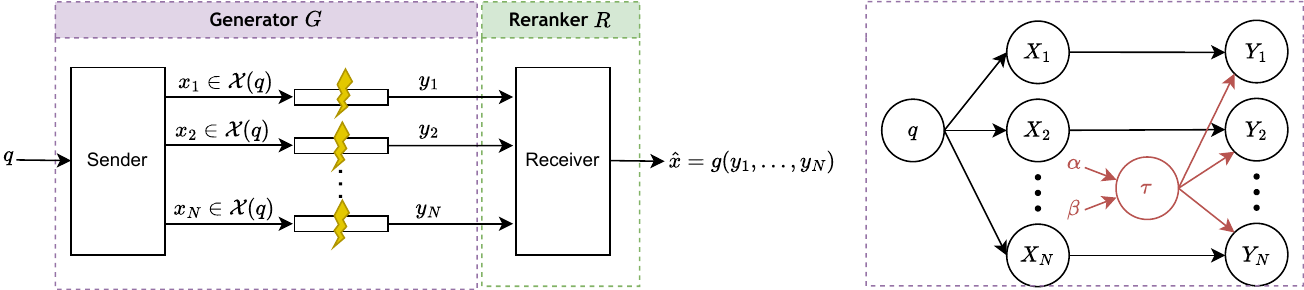}

}
\caption{\textbf{Left:} A generator-reranker system $(G, R)$ depicted as a communication system (\cref{sec:A Communication-Theoretic View of Reranking}). Given a query $q$ with acceptance set $\mathcal{X}(q)$, the sender sends $N$ descriptions through noisy channels. The receiver's goal is to decode an acceptable answer through reranking.  \textbf{Right:} Graphical model of the generator $G$. We consider two different models: a simplified version with $N$ independent hypotheses, represented in black (\cref{sec:Independent hypotheses}), and a scenario with exchangeable hypotheses, represented in red (\cref{sec:Exchangeable hypotheses}).}
\label{fig:communication-system}
\end{center}
\end{figure}

In this paper, we draw a parallel between these two worlds by regarding generator-reranker LLMs as communication systems (\cref{sec:A Communication-Theoretic View of Reranking} and \cref{fig:communication-system}, left).
We conceptualize the LLM generator $G$ as a sender transmitting $N$ message descriptions in parallel through noisy channels, leading to $N$ potentially corrupted hypotheses. Then, the receiver, which corresponds to the reranker $R$, decodes the message by ranking the potentially corrupted descriptions and selecting the one found to be most reliable.
The goal is for the combined $(G, R)$ system to have lower error rate than $G$ alone, and for the error rate to decay quickly with $N$. 
Our main contributions are as follows: 

\begin{itemize}[leftmargin=*]
    \item We show that when the channel distributions are independent, this simple protocol is asymptotically error-free (\textit{i.e.},  it generates an acceptable answer almost surely when $N \rightarrow \infty$), even in scenarios where the reranker is imperfect, \textit{e.g.}, governed by a Mallows or a Zipf-Mandelbrot model. In the former case, the error probability decays exponentially fast (\cref{sec:Independent hypotheses}).
    \item We show that the protocol is still asymptotically error-free if we assume that the channel distributions are statistically dependent. When they are coupled by a Beta prior, we show that the error probability decays as a power law when the reranker is perfect (\cref{sec:Exchangeable hypotheses}).
    \item We use our framework to obtain ``reranking laws'', which we validate  empirically on text-to-code generation with DeepSeek-Coder 7B (\cref{subsec:Code generation}), on machine translation of medical data with TowerInstruct 13B (\cref{subsec:Machine translation}), and on mathematical and commonsense reasoning benchmarks (\cref{app:additional-experiments}).
\end{itemize}

\paragraph{Notation.} We denote $[N] := \{1,...,N\}$ and we use the shorthand notation $X_{1:N} := (X_1, ..., X_N)$. We use capital letters $(X, Y, ...)$ for random variables and represent probability distributions by $\mathbb{P}(X), \mathbb{P}(Y)$, etc. We denote expectations of functions $f$ under $\mathbb{P}(X)$ by $\mathbb{E}_X[f(X)]$.

\section{A Communication-Theoretic Perspective of Generator-Reranker Systems}
\label{sec:A Communication-Theoretic View of Reranking}

The focus of our paper is on \textbf{generator-reranker systems}: a \textbf{generator} $G$ (such as an LLM) is prompted with a \textbf{query} $q$ (\textit{e.g.}, a question to be answered, a source text to be translated, or a textual prompt for code). As a response to this query, $G$ generates $N$ candidate answers $y_1, ..., y_N$ (called \textbf{hypotheses}). We are agnostic about the internals of $G$ and the way the hypotheses are generated: they could come from the same system through sampling or beam search, or they could come from an ensemble of different systems. 
These hypotheses are then processed by a \textbf{reranker} $R$, which ranks them and returns as the final output the one which is found to be the best answer. 
We are also agnostic about how $R$ is built---it could be an external system or it could be part of (or share parameters with) the generator. 
Commonly used rerankers are quality estimators \citep{fernandes-etal-2022-quality}, energy-based models \citep{bhattacharyya-etal-2021-energy}, reward models \citep{li-etal-2022-using}, and minimum Bayes risk decoders \citep{kumar-byrne-2002-minimum, eikema-aziz-2020-map, freitag2022high, shi-etal-2022-natural}. 

Regardless of specific design decisions, the goal of the generator-reranking system $(G,R)$ is to leverage the reranker $R$ to produce better answers (according to some quality metric) than the ones which would be obtained through $G$ alone (\textit{e.g.}, a single sample). 
In this paper, we show that the propensity for this combined system to generate unacceptable outputs, such as those containing critical errors or hallucinations, decays quickly enough with ${N}$ under mild assumptions on $G$ and $R$.

We draw an analogy with communication theory as follows. 
Let $\Sigma$ be an underlying alphabet and  $\Sigma^* := \bigcup_{i=0}^\infty \Sigma^i$ its Kleene closure, \textit{i.e.}, the set of strings from $\Sigma$. 
Given the query $q$, we denote by $\mathcal{X}(q) \subseteq \Sigma^*$ the set of \textbf{acceptable answers}.%
\footnote{A key difference between our framework and most lossless communication systems is that there is no need to communicate a \textit{specific} message---any answer in the equivalence class $\mathcal{X}(q)$ is acceptable, hence, if the decoder recovers any message in this set, the communication is considered successful. This is a natural conceptualization in problems involving natural language (where a paraphrase of a correct answer is still correct) or code (where multiple programs might lead to the same execution).} %
We assume the communication system depicted in \cref{fig:communication-system} (left), a form of \textbf{multiple description source coding} \citep{Ozarow1980On,Gamal1982Achievable,Laneman2005channel}. 
In this framework, the sender transmits $N$ acceptable answers (called \textbf{descriptions}) $x_1, ..., x_N \in \mathcal{X}(q)^N$ in parallel through noisy channels. These descriptions are corrupted according to a distribution $\mathbb{P}(y_1, ..., y_N | x_1, ..., x_N)$, so that some  hypotheses $y_i$ may become unacceptable ($y_i \in \Sigma^* \setminus \mathcal{X}(q)$). This ``channel noise'' is a way to conceptualize the imperfections of the generator $G$.
On the receiver side, a decoder processes the (potentially) corrupted descriptions and estimates $\hat{x} = g(y_1, ..., y_N)$ using some decoding function $g$.
The overarching goal is to achieve a low error probability $P_\mathrm{err}(N; q) := \mathbb{P}(\hat{X} \notin \mathcal{X}(q) \mid q)$ for any query $q$. 
By bounding the maximal probability of error (over all queries), the average error probability is automatically bounded \citep[\S 8]{cover2012elements}. 
In this paper, we focus on rerankers as the decoding functions, where $g(y_1, ..., y_N)$ returns the top ranked answer, \textit{i.e.}, $g(y_1, ..., y_N) = y_i$ for some $i \in [N]$.

We formalize this construction by considering different models for $G$ and $R$ in the following sections, studying the conditions under which the resulting protocol is \textbf{asymptotically error-free}:

\begin{definition}\label{def:direct_sum}
A protocol is  asymptotically error-free 
if, for any query $q$, the probability of the decoder outputting an unacceptable answer approaches zero as $N$ tends to infinity, \textit{i.e.},
\begin{equation}
    \lim_{N \rightarrow \infty} \underbrace{\mathbb{P}(g(Y_1, ..., Y_N) \notin \mathcal{X}(q) \mid q)}_{:= P_\mathrm{err}(N; q)} = 0.
\end{equation}
\end{definition}

For simplicity, we assume that $X_1, ..., X_N$ are conditionally independent given the query $q$, \textit{i.e.}, that $\mathbb{P}(x_1, ..., x_N | q) = \prod_{i=1}^N \mathbb{P}(x_i | q)$.%
\footnote{In fact, all results in this paper still hold if there are dependencies between $X_1, ..., X_N$.} %
We also assume that $Y_{1:N}$ are independent from $q$ given $X_{1:N}$ such that $q \rightarrow X_{1:N} \rightarrow Y_{1:N}$ forms a Markov chain. 
Taken together, these two assumptions mean that $\mathbb{P}(x_{1:N}, y_{1:N} | q) = \mathbb{P}(x_{1:N} | q) \mathbb{P}(y_{1:N} | x_{1:N}) = \left(\prod_{i=1}^N \mathbb{P}(x_i | q) \right) \mathbb{P}(y_{1:N} | x_{1:N}).$

\section{Generator-Reranker Systems with Independent Hypotheses}
\label{sec:Independent hypotheses}

We first consider the case where the corrupted descriptions $Y_{1:N}$ are conditionally independent and identically distributed (i.i.d.) given $X_{1:N}$ and where $Y_i$ depends only on $X_i$, that is, $\mathbb{P}(y_{1:N} | x_{1:N}) = \prod_{i=1}^N \mathbb{P}(y_i | x_i)$. 
Conceptually, this is the scenario where the parallel channels do not interfere, and it corresponds to the graphical model shown in \cref{fig:communication-system} (right) without the part in red. 
While this case may not be very realistic in practice---for example, when the hypotheses produced by the generator are all sampled from the same model---it makes the analysis simpler. We will show later in \cref{sec:Exchangeable hypotheses} how the analysis can be extended when this assumption does not hold, reusing the results from this section. 

In the sequel, given a query $q$, 
we let $\epsilon$ denote the probability of a hypothesis being unacceptable, $\epsilon := \mathbb{P}(Y_i \notin \mathcal{X}(q) \mid X_i = x_i, q) = \mathbb{P}(Y_i \notin \mathcal{X}(q) \mid X_i = x_i)$.

\subsection{Perfect and random rerankers}\label{subsec:perfect_random_rerankers}
We start by assuming that $R$ is a \textbf{perfect reranker}, which implies that it produces an acceptable output when presented with a set of $N$ hypotheses if and only if at least one of them is acceptable.
In this case, the error probability becomes
\begin{align}\label{eq:error_perfect_reranker}
    P_\mathrm{err}(N; q) = \mathbb{P}(g(Y_1, ..., Y_N) \notin \mathcal{X}(q) \mid q) &= \mathbb{E}_{X_{1:N} | q} \big[\mathbb{P}(g(Y_1, ..., Y_N) \notin \mathcal{X}(q) \mid X_{1:N}, q) \big] 
    \nonumber\\ &= \mathbb{E}_{X_{1:N} | q} \big[\mathbb{P}(Y_i \notin \mathcal{X}(q), \,\, \forall i \in [N] \mid X_{1:N})\big]
    \nonumber\\ &= \mathbb{E}_{X_{1:N}|q} \bigg[\prod_{i=1}^N \underbrace{P(Y_i \notin \mathcal{X}(q) \mid X_i)}_{= \epsilon}\bigg] =  \epsilon^N.
\end{align}
Thus, $P_\mathrm{err}(N; q)$ goes to zero exponentially fast with $N$ for any $\epsilon \in [0, 1)$, indicating that when the hypotheses are independent and the reranker is perfect, the protocol is error-free. 

On the other end of the spectrum, if the reranker is \textbf{random}---\textit{i.e.}, if it selects one of the $N$ hypotheses uniformly at random, we obtain
\begin{align}\label{eq:error_random_reranker}
    P_\mathrm{err}(N; q) = \mathbb{P}(g(Y_1, ..., Y_N) \notin \mathcal{X}(q) \mid q) &= \mathbb{E}_{X_{1:N} | q} \big[\mathbb{P}(g(Y_1, ..., Y_N) \notin \mathcal{X}(q) \mid X_{1:N}, q) \big] 
    \nonumber\\ &= \mathbb{E}_{X_{1:N} | q} \bigg[ \mathbb{E}_{i} \big[\mathbb{P}(Y_i \notin \mathcal{X}(q) \mid X_{1:N}, i)\big]\bigg] = \epsilon, 
\end{align}
that is, we obtain the same error probability as the generator alone, as expected.

\subsection{Imperfect reranker: Mallows model}\label{subsec:mallows}

We consider now more realistic rerankers. 
A statistical ranking model widely used  in machine learning applications is the \textbf{Mallows model} \citep{klementiev2008unsupervised,klementiev2009unsupervised,chierichetti2018mallows,tang2019mallows}. 
Let $\Pi$ denote the set of permutations over $N$ elements, and let $d : \Pi \times \Pi \rightarrow \mathbb{R}_+$ be a distance function between permutations. 
In this paper, we use the Kendall-tau distance $d(\pi, \pi')$, which returns the number of adjacent transpositions needed to turn $\pi$ into $\pi'$. 
Given a location parameter $\pi_0 \in \Pi$ and a scale parameter $\lambda \in \mathbb{R}_+$, the probability of a ranking $\pi$ according to the Mallows model is $\mathbb{P}(\pi; \pi_0, \lambda) = \exp(-\lambda d(\pi, \pi_0)) / Z(\lambda)$, where $Z(\lambda)$ is the partition function. 

In our setting, we assume that $\pi_0$ is the ground truth (oracle) ranking%
\footnote{More specifically, we assume that the hypotheses are ranked according to some quality metric compatible with $\mathcal{X}(q)$, that is, unacceptable answers should be ranked after acceptable answers.} %
of the hypotheses $y_1, ..., y_N$ and $\pi$ is the ranking obtained by the reranker model, so that $\mathbb{P}(\pi; \pi_0, \lambda)$ expresses how imperfect the reranker might be. Note that the family of Mallows models include both perfect and random rerankers as limit cases, respectively as $\lambda \rightarrow +\infty$ and as $\lambda = 0$.\footnote{Notably, $e^{-\lambda}$ strictly between 0 and 1 correspond to imperfect rerankers that are better than random. Lower values indicate higher-quality rerankers, making $e^{-\lambda}$ an inverse measure of reranker quality.}

\begin{figure}[t]
\begin{center}
\centerline{
\includegraphics[width=1.0\columnwidth]{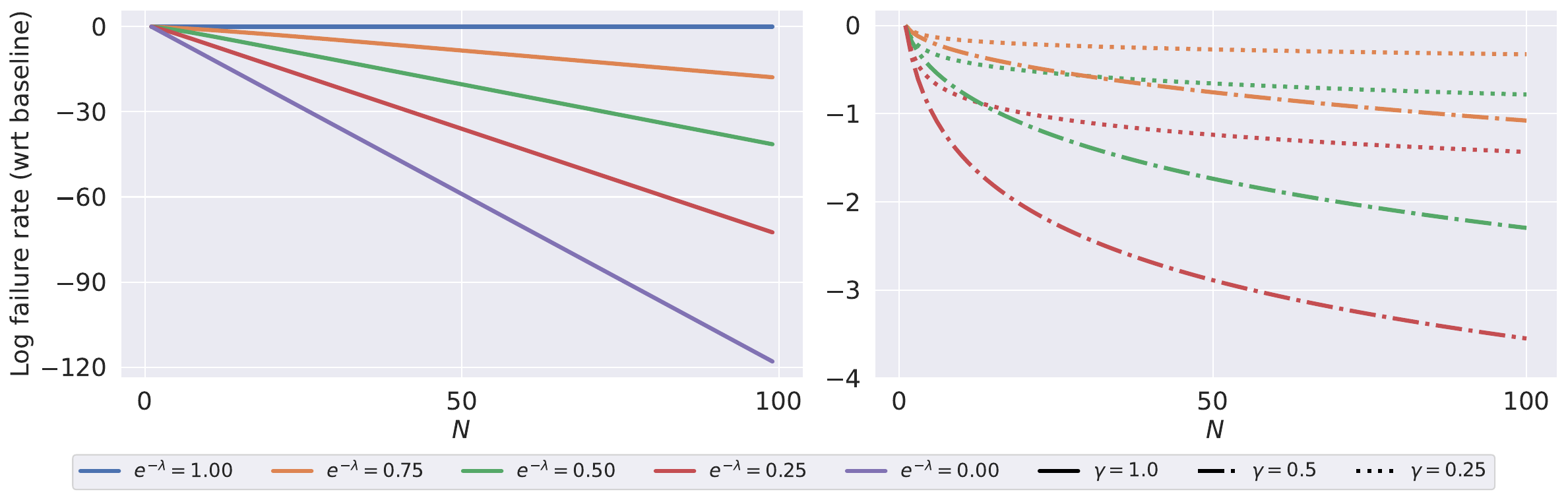}
}
\caption{Log of the failure rate (difference with respect to the baseline rate $\log \epsilon$) as a function of the number of generated independent hypotheses $N$ for several values of $e^{-\lambda}$ and $\epsilon=0.3$. \textbf{Left:} Mallows model (\cref{subsec:mallows}). \textbf{Right:} Zipf-Mandelbrot model (\cref{subsec:zipf}).}
\label{fig:reranker_independent}
\end{center}
\end{figure}

Let $\eta_j$ denote the marginal probability that the reranker places at the top the  $j\textsuperscript{th}$ highest ranked hypothesis according to the oracle, \textit{i.e.}, $\eta_j = \mathbb{P}(\pi_0(\pi^{-1}(1)) = j)$. 
When $K$ out of the $N$ hypotheses are unacceptable, the reranker will pick an unacceptable hypothesis with probability $\sum_{j = N-K+1}^N \eta_j$. Combining this with the fact that the probability of $G$ generating $K$ unacceptable hypotheses is a binomial distribution, the error probability becomes
\begin{align}\label{eq:error_imperfect_reranker}
    P_\mathrm{err}(N; q) = \mathbb{P}(g(Y_1, ..., Y_N) \notin \mathcal{X}(q) \mid q) &= \mathbb{E}_{X_{1:N} | q} \big[\mathbb{P}(g(Y_1, ..., Y_N) \notin \mathcal{X}(q) \mid X_{1:N}, q) \big]
    \nonumber\\ &= \sum_{K=0}^N \left[ {N \choose K} \epsilon^K (1-\epsilon)^{N-K} \sum_{j = N-K+1}^N \eta_j \right].
\end{align}
Note that \eqref{eq:error_imperfect_reranker} holds for \textbf{any reranker} with top-1 (marginal) probability mass function $\bm{\eta} = [\eta_1, ..., \eta_N]$, not only Mallows models. Naively determining $\bm{\eta}$ would require marginalizing $\mathbb{P}(\pi; \pi_0, \lambda)$ by summing over all permutations $\pi$ satisfying $\pi_0(\pi^{-1}(1)) = j$, which is intractable due to the factorial number of terms involved. Fortunately,  tractable combinatorial expressions exist for  Mallows models \citep{fligner1986distance,lebanon2008non}:
the partition function has the compact expression $Z(\lambda) = \prod_{j=1}^N ({1-e^{-\lambda j}})/({1-e^{-\lambda}})$,
and we have
\citep[Prop. 5]{lebanon2008non}:
\begin{align}\label{eq:softmax}
    \eta_j &=  
    Z^{-1}(\lambda) \sum_{\pi: j = \pi_0(\pi^{-1}(1))} e^{-\lambda d(\pi, \pi_0)} 
    = \frac{e^{-\lambda(j-1)}}{\sum_{r=1}^{N} e^{-\lambda(r-1)}}.
\end{align}
Plugging \eqref{eq:softmax} into \eqref{eq:error_imperfect_reranker}, invoking the binomial theorem, and simplifying, we obtain
\begin{align}\label{eq:p_bad_mallows_indep}
    P_\mathrm{err}(N; q) = \mathbb{P}(g(Y_1, ..., Y_N) \notin \mathcal{X}(q) \mid q) = 
    \left\{
    \begin{array}{ll}
        \epsilon & \text{if $\lambda=0$}\\
        \frac{[e^{-\lambda}(1-\epsilon) + \epsilon]^N - e^{-\lambda N}}{1-e^{-\lambda N}} & \text{otherwise}.
    \end{array}
    \right.
\end{align}
Notably, when $\lambda \rightarrow +\infty$ (perfect reranking), the failure probability becomes $\epsilon^N$, as expected (see \eqref{eq:error_perfect_reranker}), demonstrating the model's ability to interpolate between scenarios of random reranking ($\lambda=0$) with a failure probability of $\epsilon$ (see \eqref{eq:error_random_reranker}), and optimal reranking ($\lambda\rightarrow +\infty$) with a failure probability of $\epsilon^N$.
A plot is shown in \cref{fig:reranker_independent} (left), for several values of $e^{-\lambda} \in [0,1]$. 

Our next result, proved in \cref{proof:prop_error_free_mallows}, shows that, even with an imperfect reranker, an asymptotically error-free protocol is possible:

\begin{proposition}\label{prop:error_free_mallows}
When $R$ is a Mallows reranker, for any $\lambda > 0$, the protocol is asymptotically error-free and the error probability decays exponentially fast, $P_\mathrm{err}(N; q) = \mathcal{O}((e^{-\lambda}(1-\epsilon) + \epsilon)^N)$. 
\end{proposition}

This result shows that $P_\mathrm{err}(N; q)$ converges Q-linearly to zero with rate of convergence $e^{-\lambda}(1-\epsilon) + \epsilon > \epsilon$. 
Therefore, \textbf{Mallows rerankers behave asymptotically as a perfect reranker but where the generator has an increased error probability.} 

Given this result, one might wonder whether any reranker ``slightly better than random'' suffices to obtain an asymptotically error-free protocol. This it \textbf{not} the case, as the next counter-example shows. 

\begin{example}\label{example:polynomial}
    Assume a reranker with probability mass function $\eta_j \propto (N-j+1)$. The resulting protocol is not asymptotically error-free; we have $P_\mathrm{err}(N; q) = \mathcal{O}(\epsilon^2)$. Therefore, the error is reduced from $\mathcal{O}(\epsilon)$ to $\mathcal{O}(\epsilon^2)$ but it is not eliminated. 
    More generally, if $\eta_j \propto (N-j+1)^r$ for a fixed positive integer $r$, we have $P_\mathrm{err}(N; q) = \mathcal{O}(\epsilon^{r+1})$. 
    See \cref{proof:example_polynomial} for a proof and plots.
\end{example}

Next, we present a class of rerankers weaker than Mallows which still lead to  error-free protocols.

\subsection{Imperfect reranker: Zipf-Mandelbrot model}\label{subsec:zipf}

For Mallows models (using the Kendall-tau distance), the marginal probabilities \eqref{eq:softmax} can be written as $\bm{\eta} = \mathrm{softmax}(-\lambda \bm{z})$, where $\bm{z} = [0, 1, ..., N-1]^\top$. We now consider transformations that yield distributions with heavier tails, which we will see later in \cref{sec:Experiments} to be a better empirical fit in several applications.
A known extension to softmax is the \textbf{$\boldsymbol{\gamma}$-entmax} \citep{peters-etal-2019-sparse},\footnote{\citet{peters-etal-2019-sparse} call this $\alpha$-entmax; we use $\gamma$ instead not to clash the notation to be introduced in \cref{sec:Exchangeable hypotheses}.} a family of transformations parametrized by $\gamma\geq 0$,
\begin{equation}\label{eq:entmax}
    \text{$\gamma$-$\mathrm{entmax}$}(\boldsymbol{z}) := [1 + (\gamma-1)(\boldsymbol{z} - \tau \mathbf{1})]_+^{1/(\gamma-1)},
\end{equation}
which recovers softmax as a limit case when $\gamma \rightarrow 1$. 
In \eqref{eq:entmax}, $\tau$ is a constant which ensures that $\text{$\gamma$-$\mathrm{entmax}$}(\boldsymbol{z})$ is normalized. 
When $\gamma > 1$, $\gamma$-entmax can return sparse distributions \citep{blondel2020learning}. 
Conversely, when $\gamma<1$, $\gamma$-entmax leads to heavy-tailed distributions (see \cref{app:entmax}).

Let us now consider $\bm{\eta} = \text{$\gamma$-$\mathrm{entmax}$}(-\lambda \bm{z})$, where $\bm{z} = [0, 1, ..., N-1]^\top$, instead of \eqref{eq:softmax}. 
Letting $p := 1/(1-\gamma)$, $b = \lambda / p$, and $a = \frac{p+\tau}{\lambda}-1$ (where $a$ is seen here as a normalizing constant that replaces $\tau$), and assuming $a > -1$ and $\gamma < 1$, we can write the $\gamma$-entmax model 
as $\eta_j = {b^{-p}}{(a + j)^{-p}}$. 
Note that $\gamma < 1$ is equivalent to $p > 1$. 
This is called a \textbf{Zipf-Mandelbrot model} \citep{zipf1932selected,mandelbrot1965information}. This model generalizes the famous Zipf's law, which applies empirically to many practical contexts, such as the frequency table of words in a corpus of natural language \citep{powers1998applications}. The  constant $a$ is determined to satisfy $\sum_{j=1}^N {(a+j)^{-p}} = b^{p}$. 
When $N \rightarrow \infty$, the left hand side becomes the Hurwitz zeta function \citep{hurwitz1882einige}, which equals the Riemann's zeta when $a = 0$, 
\begin{align}\label{eq:hurwitz_zeta}
\zeta(p, a+1) := \sum_{j=1}^\infty \frac{1}{(a+j)^p} = \frac{1}{\Gamma(p)}\int_0^{\infty} dt \frac{t^{p-1}}{e^{(a+1)t}(1 - e^{-t})}. 
\end{align}
The following result, proved in \cref{proof:prop_error_free_zipf}, shows that Zipf-Mandelbrot rerankers (which are weaker than Mallows rerankers and become the latter when $\gamma \rightarrow 1$) still ensure error-free protocols. The proof makes use of the integral representation of the Hurwitz zeta function  \eqref{eq:hurwitz_zeta} and of the dominated convergence theorem, reusing the result for Mallows models in \cref{prop:error_free_mallows}.

\begin{proposition}\label{prop:error_free_zipf}
    When $R$ is a Zipf-Mandelbrot reranker, for any $\lambda > 0$ and $\gamma < 1$, the protocol is asymptotically error-free. 
\end{proposition}

\cref{fig:reranker_independent} (right) shows how this model differs from the one presented in \cref{subsec:mallows}.
Since the reranker is weaker, the error curves bend causing the error decrease to be slower, but still convergent to zero.

\section{Generator-Reranker Systems with Dependent Hypotheses}
\label{sec:Exchangeable hypotheses}

\begin{wrapfigure}[17]{r}{0.50\textwidth}
\centering
\includegraphics[width=0.5\columnwidth]{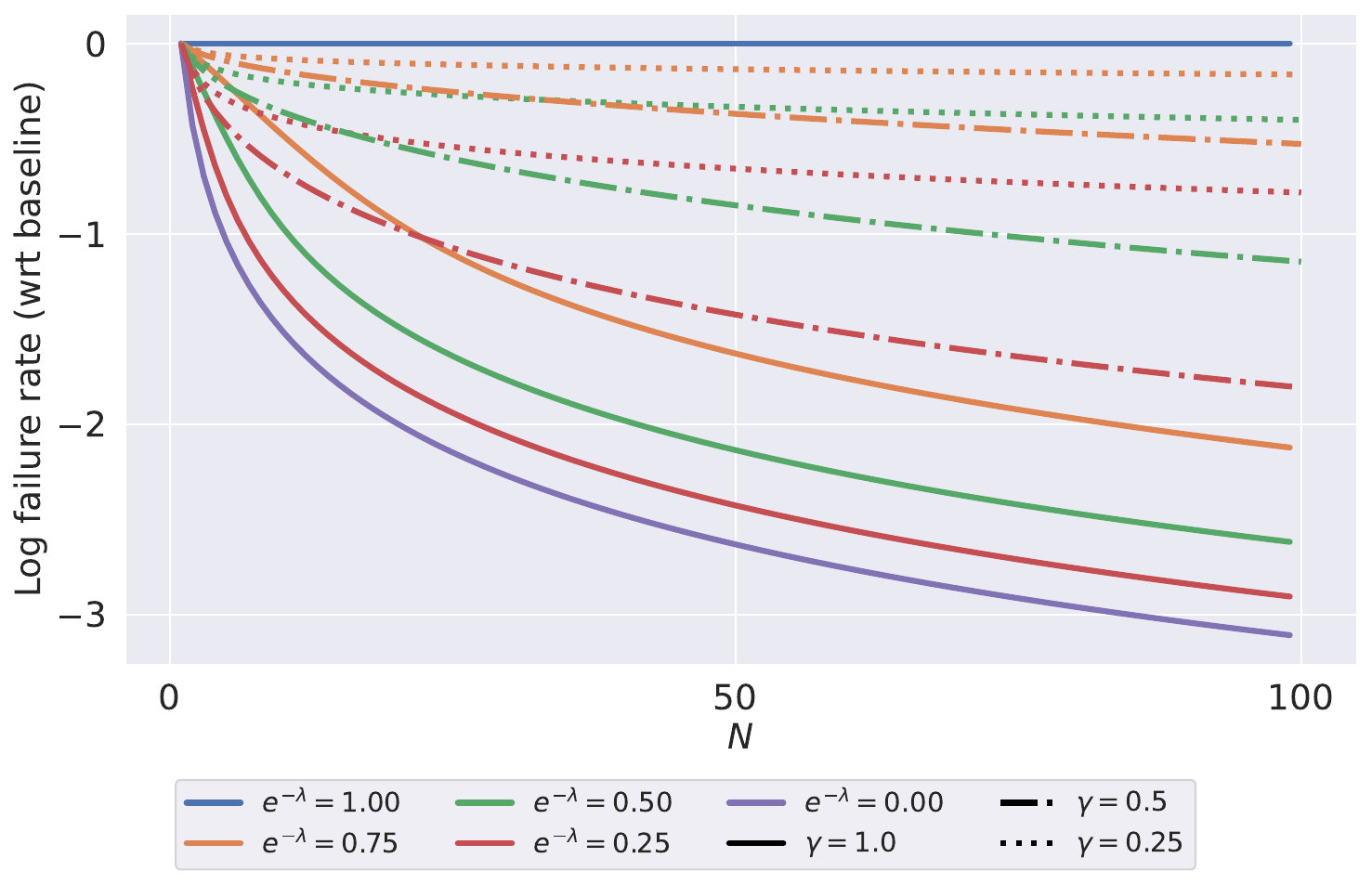}
\caption{Log of the failure rate as a function of the number of generated \textbf{exchangeable} hypotheses $N$ for several values of $\gamma$, $e^{-\lambda}$, and $\epsilon=\alpha=0.3$.}
\label{fig:reranker_exchangeable}
\end{wrapfigure}
We assume now a more realistic scenario where the independence assumption of \cref{sec:Independent hypotheses} might not hold. 
For example, $(X_1, Y_1), ..., (X_N, Y_N)$ might be only \textbf{exchangeable}---this is the case, for example, when the hypotheses are generated from $G$ by sampling from a given model, conditioned on the query. 
In communication theory parlance, this assumes the presence of channel ``interference'' that introduces dependencies between the errors at the various channels, although permuting the messages at each channel does not change the joint distribution. 
By de Finetti's theorem \citep{diaconis1980finite}, exchangeability implies that there is some mixture variable $h \in \mathcal{H}$ such that $\mathbb{P}(x_{1:N}, y_{1:N}) =
\int_\mathcal{H} d\mathbb{P}(h) \prod_{i=1}^N \mathbb{P}(x_i | h) \mathbb{P}(y_i | x_i, h)$. 

We assume further that $h = (q, \tau)$ can be decoupled into the query variable $q$, which conditions $x$, and a random variable $\tau$, which conditions $y$, such that 
$\mathbb{P}(x_i | h) := \mathbb{P}(x_i | q)$ and $\mathbb{P}(y_i | x_i, h) := \mathbb{P}(y_i | x_i, \tau)$. This corresponds to the graphical model in \cref{fig:communication-system} (right), including the part in red. 
We let $\tau$ be a continuous random variable in $[0, 1]$
such that $\mathbb{E}[\tau] = \epsilon = \mathbb{P}(Y_i \ne \mathcal{X}(q) \mid X_i)$. 
A convenient choice is a Beta distribution with parameters $\alpha$ and $\beta$,
$p(\tau; \alpha, \beta) := \frac{\Gamma(\alpha + \beta)}{\Gamma(\alpha) \Gamma(\beta)} \tau^{\alpha-1}(1-\tau)^{\beta - 1}.$

\paragraph{Perfect reranker and Beta coupling.} 
If $R$ is a perfect reranker, the error probability is
\begin{align}\label{eq:error_perfect_beta}
    P_\mathrm{err}(N; q) &= \mathbb{P}(g(Y_1, ..., Y_N) \notin \mathcal{X}(q) \mid q) = \mathbb{E}_{X_{1:N} | q} \big[\mathbb{P}(g(Y_1, ..., Y_N) \notin \mathcal{X}(q) \mid X_{1:N}\big] 
    \nonumber\\&= \mathbb{E}_{X_{1:N}} \Bigg[ \int_0^1 d\tau \,\, p(\tau) \prod_{i=1}^N \underbrace{\mathbb{P}(Y_i \notin \mathcal{X}(q) \mid X_i, \tau)}_{= \tau} \Bigg] = \mathbb{E}_\tau[\tau^N].
\end{align}
When $\tau \sim \text{Beta}(\tau; \alpha, \beta)$, the $N\textsuperscript{th}$-raw moment \eqref{eq:error_perfect_beta} has a closed form, leading to $P_\mathrm{err}(N; q) = \prod_{i=1}^{N} \frac{\alpha + i -1}{\alpha + \beta + i - 1}$. 
The next result, proved in \cref{proof:prop_error_free_dependent_perfect} using Gautschi's inequality \citep{gautschi1959some} and the Stirling's formula, shows that we still obtain an error-free protocol, albeit the error decays slower than in the independent case---no longer exponentially but rather following a power law.

\begin{proposition}\label{prop:error_free_dependent_perfect}
When $\tau \sim \text{Beta}(\tau; \alpha, \beta)$ and with a perfect reranker, the protocol is error-free and the error probability decays as a power law, $P_\mathrm{err}(N; q) = \mathcal{O}(N^{-\beta})$. Furthermore, for $\beta < 1$, we have $P_\mathrm{err}(N; q) \in \left(\frac{\Gamma(\alpha+\beta)}{\Gamma(\alpha)}(\alpha + \beta + N)^{-\beta}, \frac{\Gamma(\alpha+\beta)}{\Gamma(\alpha)}(\alpha + \beta + N - 1)^{-\beta}\right)$.
\end{proposition}

\paragraph{Imperfect reranker.}
When $\tau \sim \text{Beta}(\tau; \alpha, \beta)$, the probability of exactly $K$ out of $N$ messages being corrupted is (due to the conjugacy between the Beta prior and the binomial distribution) ${N \choose K} \int_0^1 d\tau \,\, p(\tau; \alpha, \beta)  \tau^K (1-\tau)^{N-K} =
{N \choose K} \frac{\prod_{i=1}^K (\alpha + i - 1)\prod_{i=1}^{N-K} (\beta + i - 1)}{\prod_{i=1}^{N} (\alpha + \beta + i - 1)}$.  Therefore, using the reranker marginals $\bm{\eta}$ as in \eqref{eq:error_imperfect_reranker}, we get
\begin{equation}
    P_\mathrm{err}(N; q) = \sum_{K=0}^N {N \choose K} \frac{\prod_{i=1}^K (\alpha + i - 1)\prod_{i=1}^{N-K} (\beta + i - 1)}{\prod_{i=1}^{N} (\alpha + \beta + i - 1)} \sum_{j=N-K+1}^N \eta_j,
    \label{eq:P_err(N;q)-exch-imp}
\end{equation}

which leads to the plot in \cref{fig:reranker_exchangeable} for Mallows and Zipf-Mandelbrot models.%
\footnote{Since $\tau \sim \text{Beta}(\tau; \alpha, \beta)$, we have $\mathbb{E}[\tau] = {\alpha}/({\alpha+\beta})$, which we set to $\epsilon$ to match the  independent setting from \cref{sec:Independent hypotheses}, resulting in $\beta = (\epsilon^{-1} - 1)\alpha$. 
Hence, $\alpha$ is our only new free parameter. 
As $\alpha \rightarrow 0^+$, the hypotheses become maximally dependent and reranking is hopeless; 
as $\alpha \rightarrow \infty$, the scenario reverts to full independence.}

The next result, proved in \cref{proof:prop_error_free_dependent_imperfect}, shows that the dependencies considered in this subsection do not compromise the error-free protocol when it exists for any density $p(\tau)$ which is finite in $(0,1)$ (not necessarily a Beta distribution). The proof invokes the dominated convergence theorem to enable commuting the limit with the integral sign. 

\begin{proposition}\label{prop:error_free_dependent_imperfect}
    Let $G_\tau$ be a generator producing independent hypotheses (\cref{sec:Independent hypotheses}) where each hypothesis is acceptable with probability $
    1-\tau$.  
    Let the reranker $R$ be such that $
    (G_\tau, R)$ has error probability $P_\mathrm{err}^\mathrm{indep}(N; q, \tau) \rightarrow 0$ for every $\tau \in (0,1)$ (\textit{i.e.}, it is asymptotically error-free). Assume that the function $\tau \mapsto P_\mathrm{err}^\mathrm{indep}(N; q, \tau)$ is measurable for every $N \in \mathbb{N}$. 
    Then, when $R$ is used with a generator $G$ which produces exchangeable hypotheses with arbitrary distribution $p(\tau)$, finite in $(0,1)$, the system $(G, R)$ is still asymptotically error-free.
\end{proposition}

This result has important implications: it tells us that, to design error-free protocols, it is sufficient to verify if they are error-free in the simpler case where hypotheses are independent.

\section{Experiments}
\label{sec:Experiments}
In this section, we demonstrate the validity of our reranking laws on two different tasks:\footnote{\cref{app:additional-experiments} contains additional experiments on mathematical and commonsense reasoning benchmarks.} text-to-code generation (\cref{subsec:Code generation}) and machine translation of medical data (\cref{subsec:Machine translation}). Following existing literature on scaling laws for language modeling, we fit all curves on the development set using least squares \citep[App.~E]{ghorbani2022scaling} and plot them on the \emph{unseen} test set.\footnote{We use \href{https://docs.scipy.org/doc/scipy/reference/generated/scipy.optimize.least_squares.html}{\texttt{scipy.optimize.least\_squares}}.}
In all cases, we consider the generalized model presented in \cref{sec:Exchangeable hypotheses} with parameters $\alpha$, $\beta$, and a Zipf-Mandelbrot reranker with parameters $\gamma$, and $e^{-\lambda}$, which becomes a Mallows reranker when $\gamma\rightarrow1$. 
This is done in two steps: first, we fit $\alpha$ and $\beta$ using the data for the perfect reranker ($e^{-\lambda}=0$).
Then, we fit $\gamma$ and $e^{-\lambda}$ using the already estimated $\alpha$ and $\beta$ and the data for the imperfect reranker.
Our code is available at \href{https://github.com/deep-spin/reranking-laws}{https://github.com/deep-spin/reranking-laws}.

\begin{figure}[t]
\centering
\includegraphics[width=0.49\columnwidth]{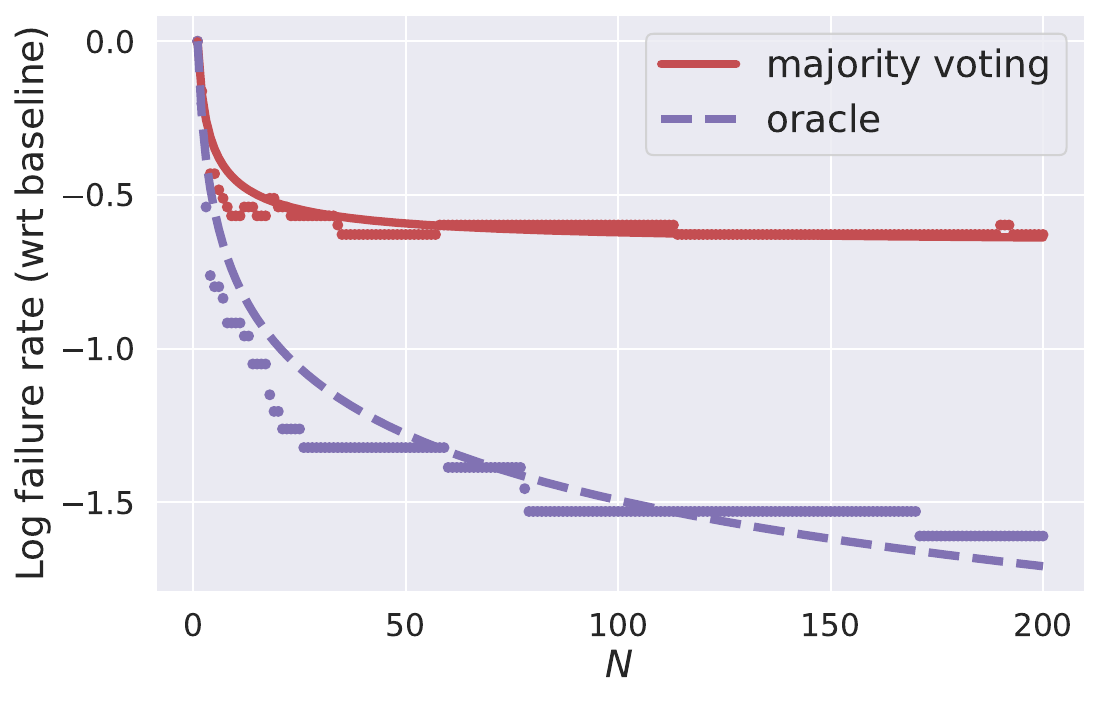}
\includegraphics[width=0.49\columnwidth]{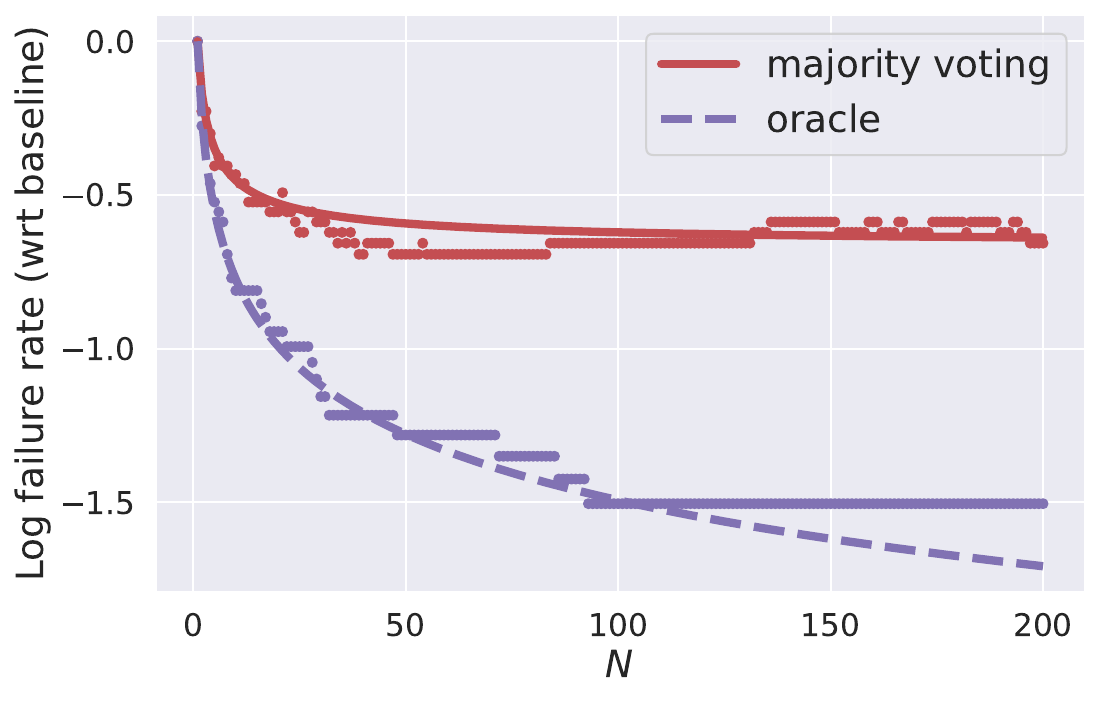}
\includegraphics[width=0.49\columnwidth]{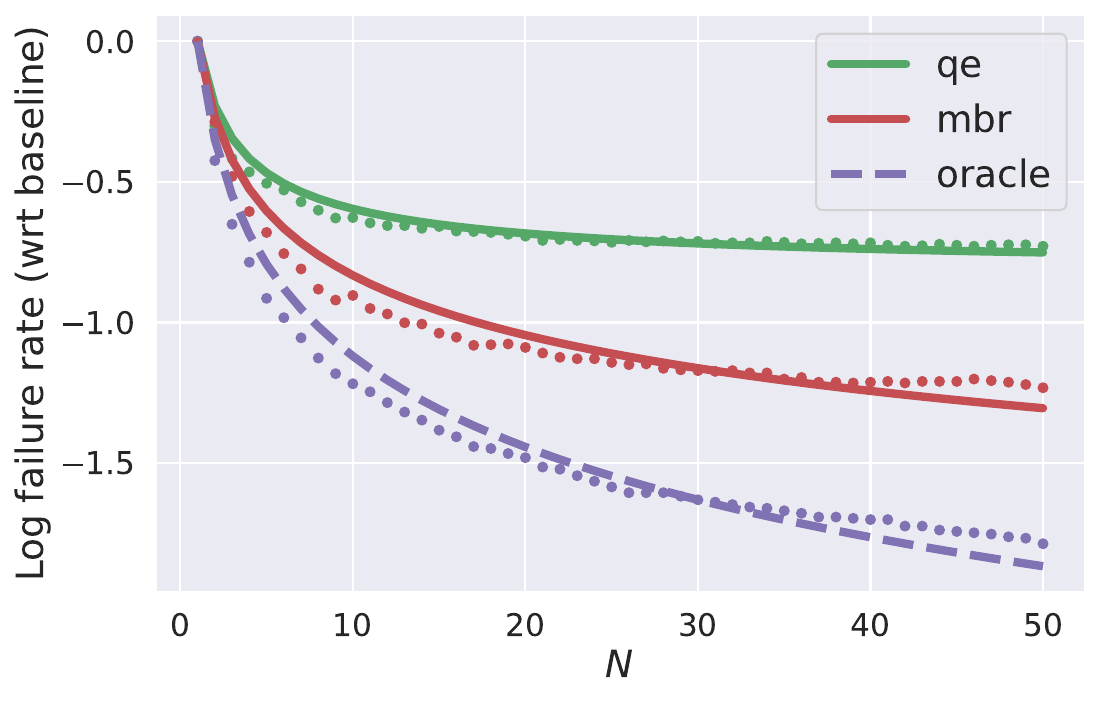}
\includegraphics[width=0.49\columnwidth]{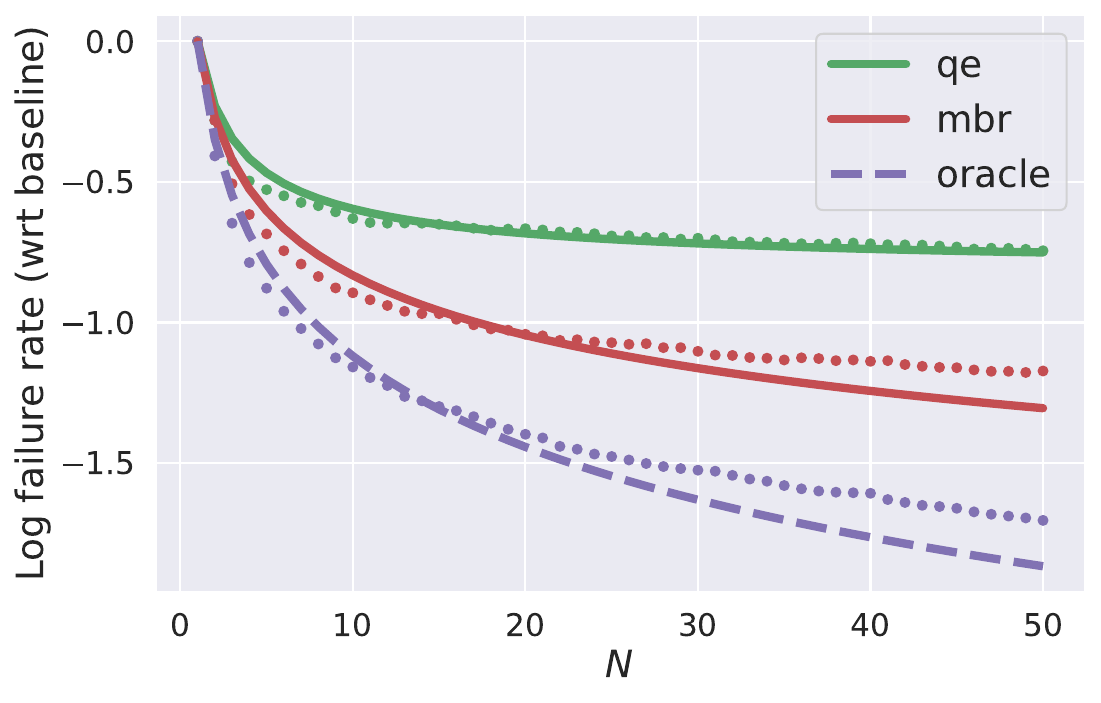}
\caption{Log of the failure rate as a function of $N$. The empirical data is represented with dots (\textbf{left:} dev, \textbf{right:} test set) and our fitted models with solid and dashed lines (imperfect and perfect reranker, respectively).
\textbf{Top:} text-to-code generation (\cref{subsec:Code generation}). \textbf{Bottom:} machine translation (\cref{subsec:Machine translation}).}
\label{fig:exps}
\end{figure}

\subsection{Code generation}
\label{subsec:Code generation}

We use a sanitized version of the MBPP dataset \citep{austin2021program, liu2023is}, a widely used benchmark for evaluating code LLMs, which includes $400$ programming problems in Python. For each problem, the dataset includes ground-truth programs and three test cases with input and ground-truth output. 
We split the dataset in two equally sized parts to get development and test splits.

We generate $200$ hypotheses with DeepSeek-Coder 7B \citep{guo2024deepseekcoder} using a sampling temperature of $1$ (see \cref{app:Text-to-code generation} for the prompt template). As in previous work, for simplicity, we use only one test case for each problem \citep{shi-etal-2022-natural}, and select one candidate by taking a \textbf{majority vote} over the execution results, dismissing hypotheses that fail to execute on the test case \citep{wang2023selfconsistency}. A hypothesis is considered unacceptable if the result of at least one test case (out of three) is different from the ground truth.

\cref{fig:exps} (top) shows the log failure rate on the dev and test sets (left and right, respectively) as a function of $N$. Even though the oracle fit is not perfect, we get $\alpha=.1$, $\beta=.309$, $\gamma=.001$, and $e^{-\lambda}=.003$ for the imperfect reranker with majority voting, which fits the data well, as shown by the red curve.

\subsection{Machine translation}
\label{subsec:Machine translation}

We use the TICO-19 dataset \citep{anastasopoulos-etal-2020-tico}, which includes 3071 English sentences in the medical domain (\textit{i.e.}, COVID-19 related content) translated into $38$ languages. We use the official splits, which contain $971$ examples for development and $2100$ for testing, focusing on translating from English (EN) to Portuguese (PT), Spanish (ES), and Russian (RU).

For each source sentence, we sample $50$ translation hypotheses with a temperature of $1$ from TowerInstruct 13B \citep{alves2024tower} using the prompt template in \cref{app:Machine translation}.\footnote{This model outperforms all existing open-source alternatives (even of larger scales) for translating content between the supported languages and is also competitive with GPT-4 \citep{openai2023gpt4}, especially when combined with MBR decoding \citep[App.~A]{alves2024tower}.}
Folowing \citet{farinhas-etal-2023-empirical}, we consider two reranking strategies: selecting the best candidate with \textbf{MBR decoding} using COMET-22 as the utility metric \citep{eikema-aziz-2020-map, rei-etal-2022-comet} and \textbf{reranking based on quality estimation} using the reference-free CometKiwi \citep{fernandes-etal-2022-quality, rei-etal-2022-cometkiwi}.
Since we cannot afford to collect human evaluation scores for each sampled hypothesis, we consider a translation to have a critical mistake (\textit{i.e.}, to be unacceptable) if its COMET-22 score is below $0.85$, and an \textbf{oracle} (perfect) reranker that picks the translation with the highest COMET-22 score.

We follow the described procedure using the data from all language pairs together. \cref{fig:exps} (bottom) shows the log failure rate on the dev and test sets as a function of $N$.
We get $\alpha=0.1$ and $\beta=0.46$. Additionally, we have $\gamma=0.182$ and $e^{-\lambda}=0.001$ for MBR decoding and $\gamma=0.001$ and $e^{-\lambda}=0.005$ for QE reranking.
See \cref{app:Machine translation} for additional plots showing these curves when the data from each language pair is used to fit a separate model. Again, we see a reasonable fit, especially for the imperfect rerankers, with MBR decoding leading to lower failure rates than reranking with QE.

\section{Discussion and Related Work}\label{sec:discussion}

We believe the communication-theoretic perspective introduced in this paper might inspire the design of new protocols for increasing the quality and safety of LLMs. 
The generator-reranker system studied in this paper bears resemblance with repetition codes, a very naive (and inefficient) class of error-correcting codes. 
Can more powerful designs \citep{hamming1950error,reed1960polynomial,gallager1962low,berrou1993near} inspire more efficient protocols? 
In machine translation, other forms of adding redundancy, such as lattice generation \citep{singhal2023eel} and hypothesis recombination \citep{vernikos2024don}, suggest that more efficient designs are indeed possible. 

Recent work also suggests that \textbf{LLM-based evaluators} could be used as highly effective rerankers in specific tasks \citep{kim2024prometheus}.
While LLMs are not yet ready to fully replace human evaluators across diverse NLP tasks \citep{bavaresco2024llmsinsteadhumanjudges}, in some cases, they can even provide fine-grained assessments
in addition to single scores \citep{kocmi-federmann-2023-gemba, fernandes-etal-2023-devil}.

Another class of communication systems allow for \textbf{feedback}, \textit{e.g.}, in ``automatic repeat request'' protocols \citep{lin1984automatic}, where the receiver has a backchannel to request the sender to retransmit missing bits of information. This framework can be useful to analyze LLM protocols where the generator generates a varying number of hypotheses interactively, relying on feedback from another module, such as a reward model or a confidence estimator, as in \citet{quach2023conformal}.  Communication with feedback was also used recently by \citet{jung2024impossible} for summarization when the generator error probability $\epsilon$ is large---our mild conditions for  asymptotically error-free protocols (Propositions~\ref{prop:error_free_mallows}--\ref{prop:error_free_dependent_imperfect}) suggest that ``bootstrapping'' a correct answer is possible even in scenarios where $G$ is very weak.
Additionally, recent work has shown that LLMs may struggle with planning or self-verification, advocating instead for tighter integration between LLMs and external model-based verifiers \citep{Kambhampati2024position}. This supports our view that using external feedback models can improve LLMs by enabling interactive, error-correcting communication.

We provide \textbf{reranking laws}, which allow us to predict how many hypotheses are necessary to achieve a desired error probability. 
This links to a rich body of literature aiming to predict the performance of deep learning models in terms of fundamental parameters, such as the model size or the amount of compute and data used to train them \citep{hestness2017deep, hestness2019beyond}. These so called ``{neural scaling laws}'' have been studied in the context of language modeling \citep{kaplan2020scaling, hoffmann2022an} and machine translation \citep{ghorbani2022scaling, fernandes2023scaling}, where we observe a power-law scaling for the performance as a function of each fundamental parameter. Our paper complements this line of work by considering the decoding dimension for generator-reranker systems. 

The analysis and theoretical results of this paper focus on binary acceptable/unacceptable decisions; however it is possible to extend our framework to consider also \textbf{continuous quality metrics} (such as \textsc{Comet} scores for translation \citep{rei-etal-2020-comet}) by replacing the notion of ``asymptotically error-free'' protocol (\cref{def:direct_sum}) by a more general concept associated to a quality target. 
A possible path is to posit a probability \textit{density} for the continuous quality metric (instead of a Bernoulli error probability) for each hypothesis coming from the generator, such as a Gaussian or uniform distribution with some input-dependent parameters. For a perfect reranker and independent hypotheses, the resulting output after reranking would be distributed according to the corresponding \textit{extreme value distribution} (this models the distribution of the \textit{highest} evaluation metric score among the $N$ hypotheses). 
Extreme value distributions are an important subject of study in order statistics \citep{david2004order} and their densities have closed form expressions in some restricted cases: for example, the Gaussian assumption above yields a Gumbel distribution, and a uniform assumption yields a Beta distribution. The asymptotic case ($N  \rightarrow \infty$) corresponds to one of Gumbel, Fréchet or Weibull families (this is a consequence of the Fisher–Tippett–Gnedenko theorem \citep{david2004order}).  From the extreme value distribution, we can obtain the \textit{expected} evaluation metric score or the probability of a quality score being below an acceptable threshold. 
However, the generalization to imperfect rerankers (such as the Mallows or Zipf-Mandelbrot rerankers described in \cref{subsec:mallows,subsec:zipf}) seems harder than in the binary case and requires further investigation.

\section{Conclusions}

We presented a communication-theoretic perspective of generator-reranker LLMs, where the generator $G$ is conceptualized as a sender transmitting $N$ descriptions in parallel through noisy channels, and the reranker $R$ decodes the message by selecting the most appropriate description. Under mild conditions, the combined system $(G, R)$ yields an acceptable answer almost surely when $N \rightarrow \infty$. Experiments on text-to-code generation and machine translation with LLMs validate our framework.

\section{Limitations and Broader Impacts}
\label{sec:Limitations and Broader Impacts}

We regard our paper as a first step connecting communication theory and LLMs, as discussed in \cref{sec:discussion}. However, it should be noted that our work has several limitations. 
First, the guarantees of error-free protocol in Propositions~\ref{prop:error_free_mallows}--\ref{prop:error_free_dependent_imperfect} are only asymptotic, and in certain cases a large $N$ may be necessary to achieve a large enough error decrease. We provide convergence rates only for Mallows rerankers (with independent hypotheses and also in the dependent case, when combined with a Beta prior). 
Second, there is no simple recipe to determine if the Mallows and Zipf-Mandelbrot reranker models are a good empirical fit to concrete rerankers. The same applies to the prior distribution $p(\tau)$ which makes hypotheses dependent. Third, while our experiments in \cref{sec:Experiments} suggest a reasonable fit in two tasks (code generation and machine translation), the fit is not perfect. A challenge is that, for large $N$, errors are rare events, and therefore prone to statistical inaccuracies (this is visible in the ``steps'' observed in the code generation plots).
Finally, although our framework focuses on binary acceptable/unacceptable decisions, it can be extended to continuous evaluation metrics, but this would require modifications to some concepts (\textit{e.g.}, the notion of asymptotically error-free protocols).
Despite these limitations, the binary case remains highly relevant in practice---for example, in code generation, where the output either executes correctly or it does not.
We expect future work to overcome some of these limitations. 

In considering the broader impact of our work, it is crucial to acknowledge its early stage and predominantly theoretical nature, which lends the discussion a speculative quality.
We believe that our research can significantly enhance the reliability of LLMs by facilitating the identification of potential system failures,
holding promise in fields such as natural language processing and computer vision, where robustness and error prediction are paramount.
While not directly addressing environmental concerns shared across different LLMs \citep{strubell-etal-2019-energy}, our work could indirectly contribute to energy efficiency efforts by quantifying the efficiency of reranking methods, potentially reducing computational requirements while maintaining requisite quality thresholds during inference. 

\section*{Acknowledgments}

We would like to thank Ben Peters, Duarte Alves, Marcos Treviso, Mário Figueiredo, Sweta Agrawal, and the SARDINE lab team for helpful discussions.
This work was supported by EU's Horizon Europe Research and Innovation Actions (UTTER, contract 101070631), by the project DECOLLAGE (ERC-2022-CoG 101088763), by the Portuguese Recovery and Resilience Plan through project C645008882-00000055 (Center for Responsible AI), and by Fundação para a Ciência e Tecnologia through contract UIDB/50008/2020.

\bibliography{custom}
\bibliographystyle{plainnat}

\appendix

\newpage

\section{Proofs and Visualizations}
\label{app:Proofs}

\subsection{Proof of \cref{prop:error_free_mallows}}\label{proof:prop_error_free_mallows}

Let $\lambda_\epsilon := -\log \left( e^{-\lambda}(1-\epsilon) + \epsilon \right)$ and define $F(N) = \log P_\mathrm{err}(N; q) = \frac{e^{-\lambda_\epsilon N} - e^{-\lambda N}}{1-e^{-\lambda N}}$. Observe that $0 < \lambda_\epsilon < \lambda$ 
for any $\lambda > 0$ and $\epsilon \in (0,1)$. 
We extend the domain of $F$ to the real numbers in $[1, +\infty)$. We will prove that $F(N)$ is decreasing and that $\lim_{N \rightarrow \infty} F'(N) = -\lambda_\epsilon$. This shows that $P_\mathrm{err}(N; q) \rightarrow 0$ at asymptotic rate $e^{{-\lambda_\epsilon}}$. 
We have $$F'(N) = \frac{(e^{-\lambda_\epsilon N} - e^{-\lambda N})'}{e^{-\lambda_\epsilon N} - e^{-\lambda N}} - \frac{(1-e^{-\lambda N})'}{1-e^{-\lambda N}} = \frac{-\lambda_\epsilon e^{-\lambda_\epsilon N} + \lambda e^{-\lambda N}}{e^{-\lambda_\epsilon N} - e^{-\lambda N}} - \frac{\lambda e^{-\lambda N}}{1-e^{-\lambda N}} \le 0,$$ hence $F(N)$ is decreasing.  
Since the second term tends to zero, the limit is given by the first term:
$$\lim_{N \rightarrow \infty} F'(N) = \lim_{N \rightarrow \infty} \frac{-\lambda_\epsilon e^{-\lambda_\epsilon N} + \lambda e^{-\lambda N}}{e^{-\lambda_\epsilon N} - e^{-\lambda N}} = \lim_{N \rightarrow \infty} \frac{-\lambda_\epsilon}{1 - e^{(-\lambda + \lambda_\epsilon)N}} + \frac{\lambda}{e^{(-\lambda_\epsilon + \lambda)N} - 1} = -\lambda_\epsilon,$$ where we used the fact that  $e^{(-\lambda + \lambda_\epsilon)N} \rightarrow 0$ and $e^{(-\lambda_\epsilon + \lambda)N} \rightarrow +\infty$. 
This proves the desired claim, that is, the error probability decreases exponentially fast with rate $e^{-\lambda_\epsilon}$. 
Note that, for a perfect reranker ($\lambda \rightarrow \infty$), we get $e^{-\lambda_\epsilon} = \epsilon$ and we recover the rate $\epsilon^N$ seen in \cref{subsec:perfect_random_rerankers}.

\subsection{Proof of \cref{example:polynomial}}\label{proof:example_polynomial}

We first provide a proof for $r=1$. We have $\sum_{j={N-K+1}}^N \eta_j = \sum_{j=1}^K \eta_{N-K+j} = \frac{\sum_{j=1}^K j}{\sum_{j=1}^N j} = \frac{K(K+1)}{N(N+1)}$. 
Plugging this into \cref{eq:error_imperfect_reranker}, we obtain 
\begin{align}
P_\mathrm{err}(N; q) &= \sum_{K=0}^N {N \choose K} \epsilon^K (1-\epsilon)^{N-K} \frac{K^2 + K}{N^2 + N} = \frac{\mathbb{E}_{K \sim B(N, \epsilon)}[K^2 + K]}{N^2 + N} \nonumber\\
&= \frac{N\epsilon(1-\epsilon) +  N^2\epsilon^2 + N\epsilon}{N(N+1)} = \frac{\epsilon(1-\epsilon) + N\epsilon^2 + \epsilon}{N+1},
\end{align}
where $B(N, \epsilon)$ denotes the binomial distribution with parameters $N$ and $\epsilon$ and we use the facts that $\mathbb{E}_{K \sim B(N, \epsilon)}[K] = N\epsilon$ and $\mathbb{E}_{K \sim B(N, \epsilon)}[K^2] = N\epsilon(1-\epsilon) +  N^2\epsilon^2$. 
Therefore, we obtain $\lim_{N\rightarrow \infty} P_\mathrm{err}(N; q) = \epsilon^2$. 

\begin{figure}[t]
\begin{center}
\centerline{
\includegraphics[width=1.0\columnwidth]{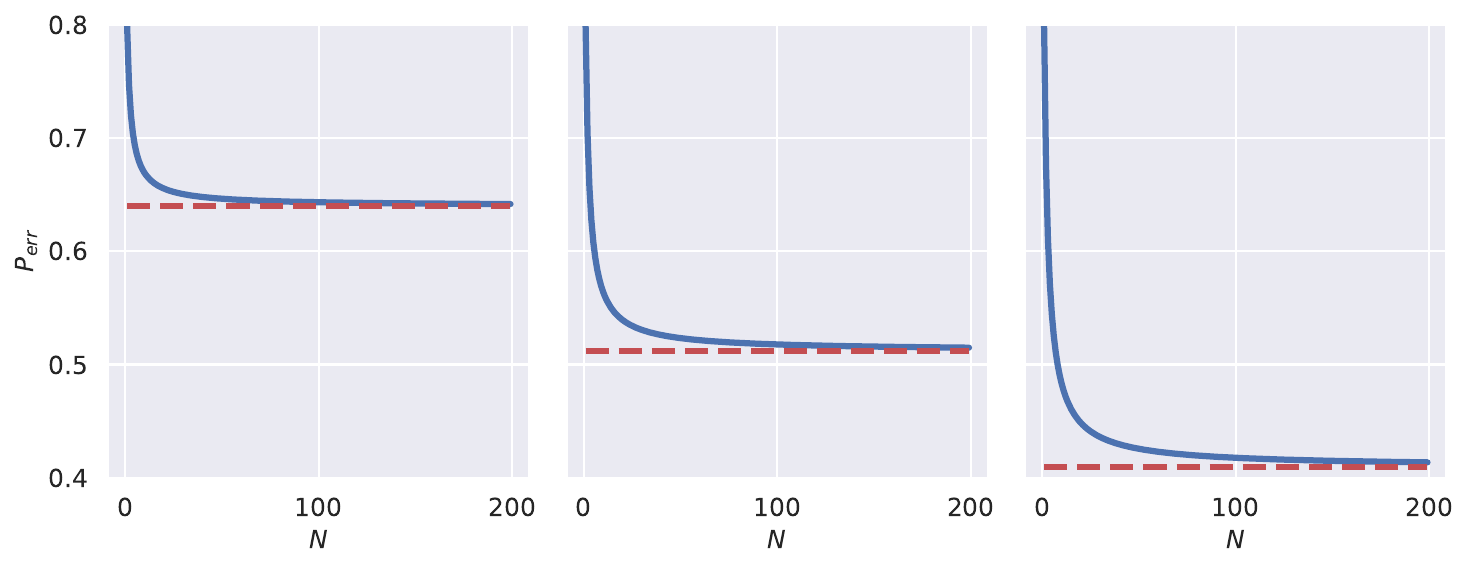}
}
\caption{$P_\mathrm{err}$ using rerankers with probability mass function $\eta_j \propto (N-j+1)^r$ with $r=\{1,2,3\}$ (from left to right) and $\epsilon=0.8$. The resulting protocol is not asymptotically error-free: the horizontal asymptotes in red correspond to $\epsilon^{r+1}$, according to \cref{eq:asymptote}.}
\label{fig:imperfect-fail}
\end{center}
\end{figure}

We now prove the general case $r \ge 1$. 
From Faulhaber's formula, we have 
$\sum_{j=1}^K j^r = \frac{1}{r+1} \sum_{j=0}^r {r+1 \choose j} B_j K^{r-j+1}$, where $B_j = \sum_{\ell=0}^j \frac{1}{\ell + 1} \sum_{m=0}^\ell {\ell \choose m} (-1)^m (m+1)^j$ denotes the $j\textsuperscript{th}$ Bernoulli number. 
Therefore, we get 
\begin{align}
\sum_{j={N-K+1}}^N \eta_j = \sum_{j=1}^K \eta_{N-K+j}^r = \frac{\sum_{j=1}^K j^r}{\sum_{j=1}^N j^r} = \frac{\sum_{j=0}^r {r+1 \choose j} B_j K^{r-j+1}}{\sum_{j=0}^r {r+1 \choose j} B_j N^{r-j+1}}. 
\end{align}
Plugging this into \cref{eq:error_imperfect_reranker}, we obtain 
\begin{align}\label{eq:polynomial_proof}
P_\mathrm{err}(N; q) &= \sum_{K=0}^N {N \choose K} \epsilon^K (1-\epsilon)^{N-K} \frac{\sum_{j=0}^r {r+1 \choose j} B_j K^{r-j+1}}{\sum_{j=0}^r {r+1 \choose j} B_j N^{r-j+1}} \nonumber\\
&= \frac{\sum_{j=0}^r {r+1 \choose j} B_j \mathbb{E}_{K \sim B(N, \epsilon)}[K^{r-j+1}]}{\sum_{j=0}^r {r+1 \choose j} B_j N^{r-j+1}}.
\end{align}
We now use the fact that the raw moments of the binomial distribution $B(N, \epsilon)$ are given by 
$\mathbb{E}_{K \sim B(N, \epsilon)}[K^m] = \sum_{\ell=0}^m {m \brace \ell} N^{\underline{\ell}} \epsilon^\ell$, 
where ${m \brace \ell} := \frac{1}{\ell!}\sum_{i=0}^{\ell}  (-1)^{\ell - i}{\ell \choose i} i^m$ are the Stirling numbers of the second kind, and $N^{\underline{\ell}} := \frac{N!}{(N - \ell)!}$ is the $\ell\textsuperscript{th}$ falling power of $N$.
Therefore, when $N \rightarrow \infty$, \eqref{eq:polynomial_proof} becomes
\begin{align}
    \lim_{N \rightarrow \infty} P_\mathrm{err}(N; q) = \lim_{N \rightarrow \infty} \frac{{r+1 \choose 0} B_0 \overbrace{{r+1 \brace r+1}}^{=1} N^{\underline{r+1}} \epsilon^{r+1}}{{r+1 \choose 0} B_0 N^{r+1}} = \epsilon^{r+1}.
    \label{eq:asymptote}
\end{align}
The plots in \cref{fig:imperfect-fail} show examples for several values of $r$. 

\begin{wrapfigure}[15]{r}{0.5\textwidth}
\vspace{-1.4em}
\includegraphics[width=0.5\columnwidth]{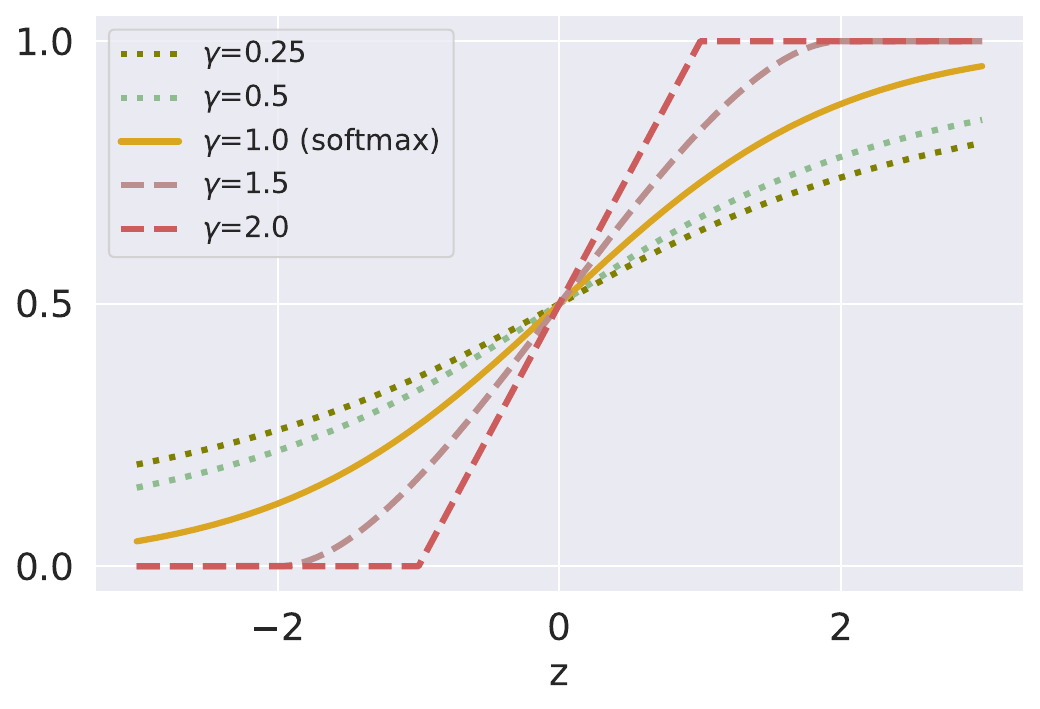}
\caption{Two-dimensional $\gamma$-entmax$([z,0])_1$.
}
\label{fig:entmax}
\end{wrapfigure}
\subsection{Entmax}
\label{app:entmax}

When $\gamma > 1$, $\gamma$-entmax can return sparse distributions \citep{blondel2020learning}. 
This case has been extensively studied as a way to, \textit{e.g.}, filter large output spaces \citep{correia2020efficient, peters-martins-2021-smoothing} or to produce more interpretable predictions \citep{correia-etal-2019-adaptively, martins2020sparse, martins2021tvmax, martins2022sparse}. 
Conversely, when $\gamma<1$, $\gamma$-entmax leads to distributions with heavier tails, which is the case of our interest, as described in \cref{subsec:zipf}. See \cref{fig:entmax} for an illustration of $\gamma$-entmax for different values of $\gamma$ in the two-dimensional case.
For $\gamma>1$, all mappings saturate at $z=\pm1/{\gamma-1}$; this does not happen for $\gamma\leq1$.

\subsection{Proof of \cref{prop:error_free_zipf}}\label{proof:prop_error_free_zipf}

Note that we can write $$\sum_{j=1}^N \frac{1}{(a+j)^p} = \zeta(p, a+1) - \zeta(p, a+N+1)$$ and 
\begin{align}
\sum_{j = N-K+1}^N \eta_j &= b^{-p}(\zeta(p, a+1) - \zeta(p, a+N+1) - \zeta(p, a+1) + \zeta(p, a+N-K+1)) \nonumber\\
&= b^{-p}(\zeta(p, a+N-K+1) - \zeta(p, a+N+1))\nonumber\\
&= \frac{1}{b^p \Gamma(p)} \int_0^{\infty} dt \frac{t^{p-1}}{e^{(a+N+1)t}(1 - e^{-t})}(e^{Kt} - 1).
\end{align}
The error probability is 
\begin{align}
P_\mathrm{err}(N; q) &= \sum_{K=0}^N \left[ \left(\begin{array}{c}N\\K\end{array}\right) \epsilon^K (1-\epsilon)^{N-K} \sum_{j = N-K+1}^N \eta_j \right] \nonumber\\
&= \frac{1}{b^p \Gamma(p)} \int_0^{\infty} dt \frac{t^{p-1}}{e^{(a+N+1)t}(1 - e^{-t})}\underbrace{\sum_{K=0}^N \left(\begin{array}{c}N\\K\end{array}\right) \epsilon^K (1-\epsilon)^{N-K} (e^{Kt} - 1)}_{=(e^t \epsilon + 1 - \epsilon)^N - 1}\nonumber\\
&= \frac{1}{b^p \Gamma(p)} \int_0^{\infty} dt \frac{t^{p-1}[(e^t \epsilon + 1 - \epsilon)^N - 1]}{e^{(a+N+1)t}(1 - e^{-t})} \nonumber\\
&= \frac{1}{b^p \Gamma(p)} \int_0^{\infty} dt \frac{t^{p-1}}{e^{(a+1)t}(1 - e^{-t})} \frac{(e^t \epsilon + 1 - \epsilon)^N - 1}{e^{tN}}\nonumber\\
&= \frac{1}{b^p \Gamma(p)} \int_0^{\infty} dt \frac{t^{p-1}}{e^{(a+1)t}(1 - e^{-t})} \underbrace{[((1 - \epsilon)e^{-t} + \epsilon)^N - e^{-tN}]}_{:= f_N(t) \rightarrow 0}.
\end{align}
Since $a$ is the normalizing constant 
such that $1 = \zeta(p, a+1) = \frac{1}{b^p \Gamma(p)} \int_0^{\infty} dt \frac{t^{p-1}}{e^{(a+1)t}(1 - e^{-t})}$ (cf. \cref{eq:hurwitz_zeta}), we can interpret the expression above as the expectation of $f_N(t) := ((1 - \epsilon)e^{-t} + \epsilon)^N - e^{-tN}$ under the probability distribution on $(0,\infty)$ with density $\pi(t) := \frac{1}{b^p \Gamma(p)} \frac{t^{p-1}}{e^{(a+1)t}(1 - e^{-t})}$. Since $f_N(t) \rightarrow 0$ pointwise for $t \in ]0,\infty[$ and it is bounded in that interval, we can 
invoke the dominated convergence theorem to commute the limit and integral sign. We then have that $P_{\mathrm{err}}(N; q) \rightarrow 0$.

\subsection{Proof of \cref{prop:error_free_dependent_perfect}}\label{proof:prop_error_free_dependent_perfect}

Let us consider first the case where $\beta=1$. Then, $$P_\mathrm{err}(N; q) = \prod_{i=1}^N \frac{\alpha + i - 1}{\alpha + \beta + i - 1} = \frac{\alpha}{\alpha + 1} \frac{\alpha + 1}{\alpha + 2} \cdots \frac{\alpha + N - 1}{\alpha + N} = \frac{\alpha}{\alpha + N} \rightarrow 0.$$ 
Now consider the case where $\beta>1$. We have for each term in the product that $\frac{\alpha + i - 1}{\alpha + \beta + i - 1} < \frac{\alpha + i - 1}{\alpha + i}$, hence we must have $P_\mathrm{err}(N; q) < \frac{\alpha}{\alpha + N}$. Since the sequence is positive (since all terms are positive) and decreasing (since all terms are $<1$), we must also have $P_\mathrm{err}(N; q) \rightarrow 0$ when $\beta > 1$. 

Finally, let us analyze the case where $0 < \beta < 1$. 
From \eqref{eq:error_perfect_beta}, we have 
\begin{align}
P_\mathrm{err}(N; q) &= \mathbb{E}_\tau[\tau^N] = \int_{0}^1 \frac{\Gamma(\alpha + \beta)}{\Gamma(\alpha) \Gamma(\beta)} \tau^{\alpha-1}(1 - \tau)^{\beta - 1} \tau^N \nonumber\\ &= 
\frac{\Gamma(\alpha + \beta)}{\Gamma(\alpha)} \frac{\Gamma(\alpha + N)}{ \Gamma(\alpha + \beta + N)}
\underbrace{\int_{0}^1 \frac{\Gamma(\alpha + \beta + N)}{\Gamma(\alpha + N) \Gamma(\beta)} \tau^{\alpha+N-1}(1 - \tau)^{\beta - 1}}_{=1} \nonumber\\
&= \frac{\Gamma(\alpha + \beta)}{\Gamma(\alpha)} \frac{\Gamma(\alpha + N)}{ \Gamma(\alpha + \beta + N)}.
\end{align}
We invoke Gautschi's inequality, which states that $x^{1-s} < \frac{\Gamma(x+1)}{\Gamma(x+s)} < (x+1)^{1-s}$ for any $x$ and $s \in (0,1)$. We set $s := 1-\beta$ and $x := \alpha + \beta + N - 1$, from which we obtain the desired result. 

To show that the error probability decays as a power law for any $\beta > 0$, we use Stirling's formula, which states that \begin{align}
    \Gamma (z)={\sqrt {\frac {2\pi }{z}}}\,{\left({\frac {z}{e}}\right)}^{z}\left(1+\mathcal{O}\left({\frac {1}{z}}\right)\right).
\end{align} 
Therefore, 
\begin{align}
    \lim_{N \rightarrow \infty}\frac{\Gamma(\alpha + N)}{\Gamma(\alpha + \beta + N)} &= \lim_{N \rightarrow \infty} \frac{{\sqrt {\frac {2\pi }{\alpha + N}}}\,{\left({\frac {\alpha + N}{e}}\right)}^{\alpha + N}}{{\sqrt {\frac {2\pi }{\alpha + \beta + N}}}\,{\left({\frac {\alpha + \beta + N}{e}}\right)}^{\alpha + \beta + N}}\nonumber\\
    &= \lim_{N \rightarrow \infty} \underbrace{\sqrt{\frac{\alpha + \beta + N}{\alpha + N}}}_{\rightarrow 1}\underbrace{{\left({\frac {\alpha + N}{\alpha + \beta + N}}\right)}^{\alpha + N}}_{\rightarrow e^{-\beta}}{{\left({\frac {\alpha + \beta + N}{e}}\right)}^{-\beta}}\nonumber\\
    &= \lim_{N \rightarrow \infty} (\alpha + \beta + N)^{-\beta} = \mathcal{O}(N^{-\beta}).
\end{align}

\subsection{Proof of \cref{prop:error_free_dependent_imperfect}}\label{proof:prop_error_free_dependent_imperfect}

Let $P_\mathrm{err}^\mathrm{indep}(N; q, \tau)$ denote the error probability of the generator-reranker system when the hypotheses are independent and each hypothesis produced by $G$ has error probability $\tau$.  
The error probability of the  generator-reranker system with exchangeable hypotheses is given by 
\begin{align}
    P_\mathrm{err}(N; q) &= \mathbb{P}(g(Y_1, ..., Y_N) \notin \mathcal{X}(q) \mid q) = \mathbb{E}_{X_{1:N} | q} \big[\mathbb{P}(g(Y_1, ..., Y_N) \notin \mathcal{X}(q) \mid X_{1:N}, q) \big]
    \nonumber\\ &= \mathbb{E}_{X_{1:N} | q} \Bigg[ \int_0^1 d\tau \,\, p(\tau) \,\, \mathbb{P}(g(Y_1, ..., Y_N) \notin \mathcal{X}(q) \mid X_{1:N}, \tau) \Bigg]
    \nonumber\\ &= \int_0^1 d\tau \,\, p(\tau) \,\, \underbrace{\mathbb{E}_{X_{1:N} | q} \Bigg[ \mathbb{P}(g(Y_1, ..., Y_N) \notin \mathcal{X}(q) \mid X_{1:N}, \tau) \Bigg]}_{= P_\mathrm{err}^\mathrm{indep}(N; q, \tau)}.
\end{align}
Therefore, $\lim_{N \rightarrow \infty} P_\mathrm{err}(N; q) = \lim_{N \rightarrow \infty} \int_0^1 d\tau \,\, p(\tau) P_\mathrm{err}^\mathrm{indep}(N; q, \tau)$. 
Since $P_\mathrm{err}^\mathrm{indep}(N; q, \tau) \in [0,1]$ for any $N \in \mathbb{N}$ and $\tau \in [0,1]$, we have that $p(\tau) P_\mathrm{err}^\mathrm{indep}(N; q, \tau) \in [0, p(\tau)]$, and therefore the integrand is bounded by $p(\tau)$, which integrates to $1$. Therefore we can invoke the dominated convergence theorem and switch the limit and integral signs. Since by assumption $\lim_{N\rightarrow \infty} P_\mathrm{err}^\mathrm{indep}(N; q, \tau) = 0$ pointwise for any $\tau \in (0,1)$, we obtain $\lim_{N \rightarrow \infty} P_\mathrm{err}(N; q) =  \int_0^1 d\tau \,\, p(\tau) \lim_{N \rightarrow \infty} P_\mathrm{err}^\mathrm{indep}(N; q, \tau) = 0$. 

\section{Experimental Details}
\label{app:Experimental Details}

\subsection{Text-to-code generation}
\label{app:Text-to-code generation}

\paragraph{Licenses.} We use DeepSeek-Coder 7B \citep{guo2024deepseekcoder}, which is available under a permissive \href{https://github.com/deepseek-ai/DeepSeek-Coder/blob/main/LICENSE-MODEL}{license} that allows for both research and unrestricted commercial use. We report results on the MBPP dataset \citep{austin2021program, liu2023is}, released under an Apache \href{https://github.com/evalplus/mbppplus_release}{license}.

\paragraph{Prompt template.}
We generate hypotheses with DeepSeek-Coder 7B \citep{guo2024deepseekcoder} using the following prompt template:
\begin{quote}

You are an AI programming assistant, utilizing the DeepSeek Coder model, developed by DeepSeek Company, and you only answer questions related to computer science. For politically sensitive questions, security and privacy issues, and other non-computer science questions, you will refuse to answer.

\#\#\# Instruction:

Please complete the following Python function in a markdown style code block:

$^{\backprime\backprime\backprime}$python

\texttt{[prompt]}

$^{\backprime\backprime\backprime}$

\#\#\# Response:

$^{\backprime\backprime\backprime}$python

\end{quote}

\paragraph{MBR-exec.}
We use MBR-exec, an approach proposed by \citet{shi-etal-2022-natural} that consists of \emph{(1)} sampling programs from an LLM, \emph{(2)} executing each program on one test case, and \emph{(3)} selecting the example with the minimal execution result-based Bayes risk. We use a 0/1 matching loss between execution results, and the Bayes risk of a program is defined by the sum of the loss between itself and the other sampled programs (the ground-truth program output is not used). We break ties by selecting the program with the smallest sampling index, corresponding to a random selection.
See \citet[Section~3]{shi-etal-2022-natural} for more details. 

\subsection{Machine translation}
\label{app:Machine translation}

\paragraph{Licenses.} We use TowerInstruct 13B \citep{alves2024tower}, which is released under a CC-BY-NC-4.0 license. We report results on the TICO-19 dataset \citep{anastasopoulos-etal-2020-tico}, publicly available through a Creative Commons CC0 \href{https://tico-19.github.io/LICENSE.md}{license}.

\paragraph{Prompt template.}
We generate hypotheses with TowerInstruct 13B \citep{alves2024tower} using the following prompt template:
\begin{quote}
\texttt{<|im\_start|>user}

Translate the following \texttt{[source language]} source text to \texttt{[target language]}:

\texttt{[source language]}: \texttt{[source sentence]}

\texttt{[target language]}: \texttt{<|im\_end|>}

\texttt{<|im\_start|>assistant}

\end{quote}

\begin{figure}[t]
\centering
\includegraphics[width=0.45\textwidth]{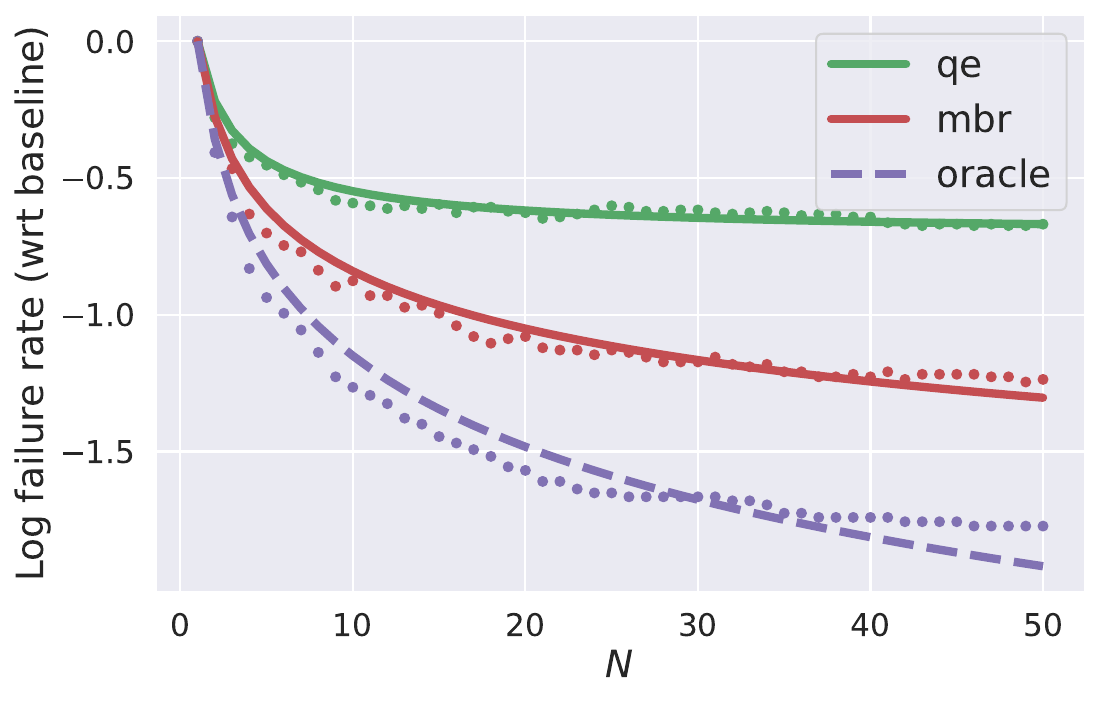}
\includegraphics[width=0.45\textwidth]{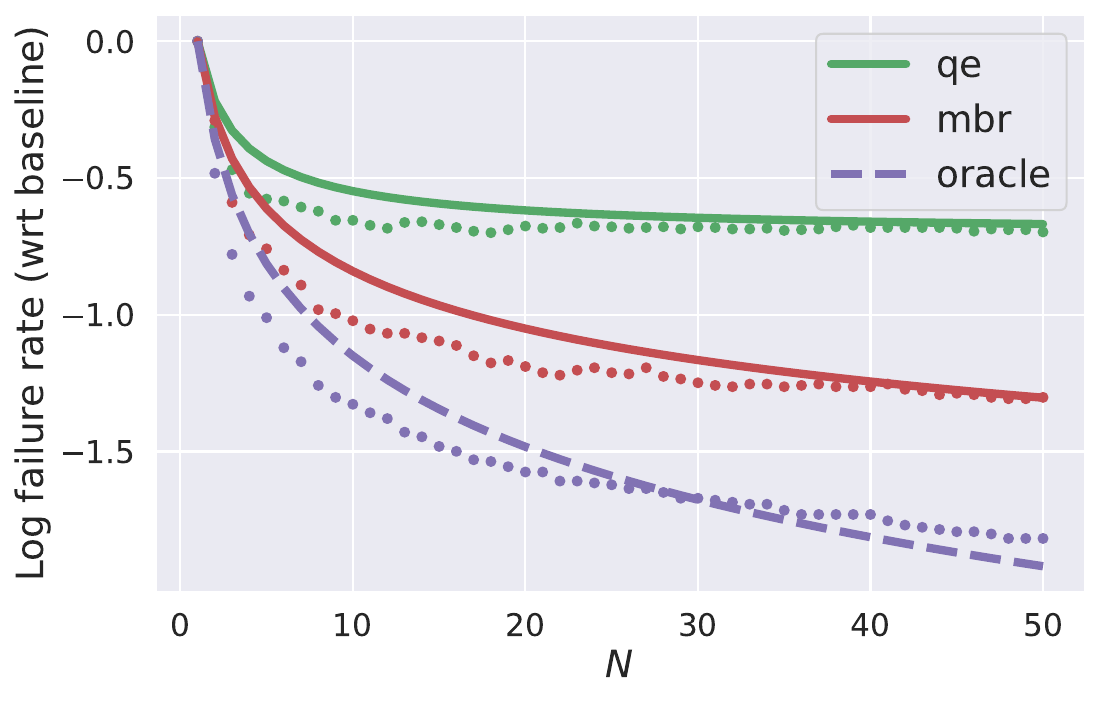}
\includegraphics[width=0.45\textwidth]{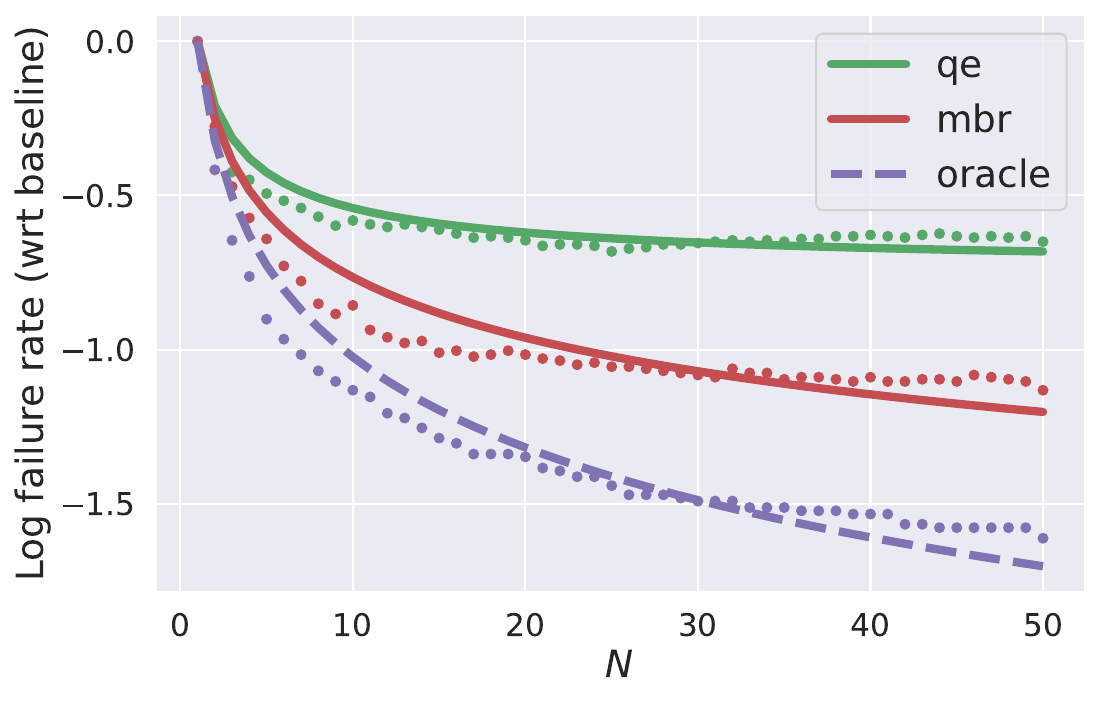}
\includegraphics[width=0.45\textwidth]{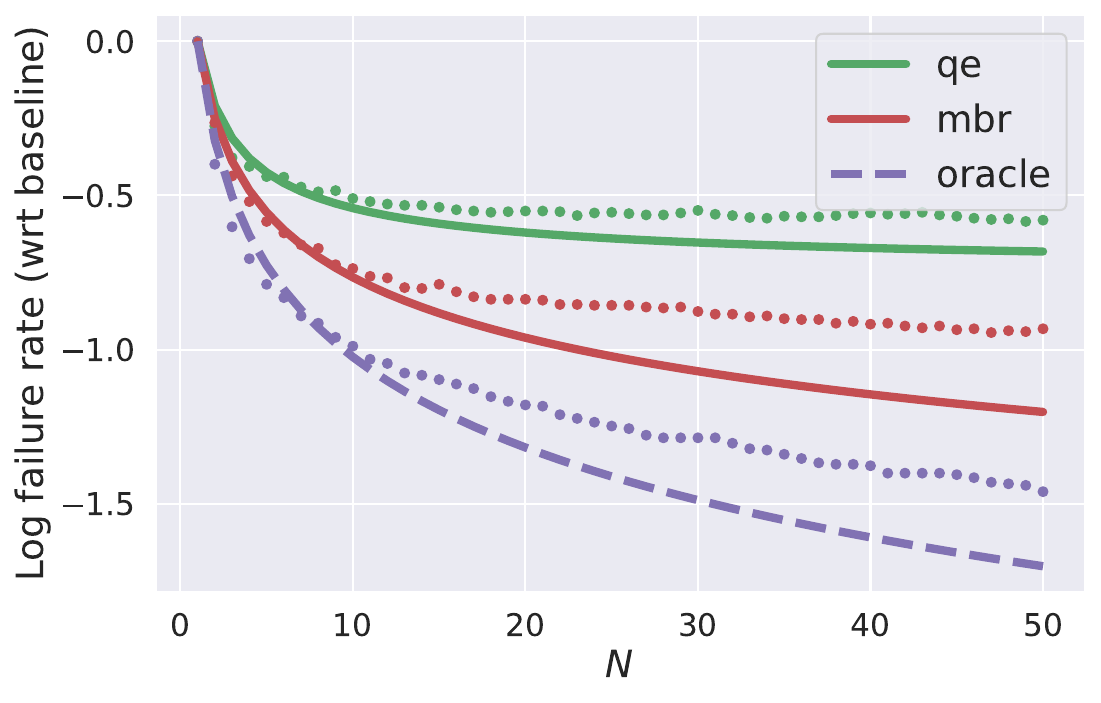}
\includegraphics[width=0.45\textwidth]{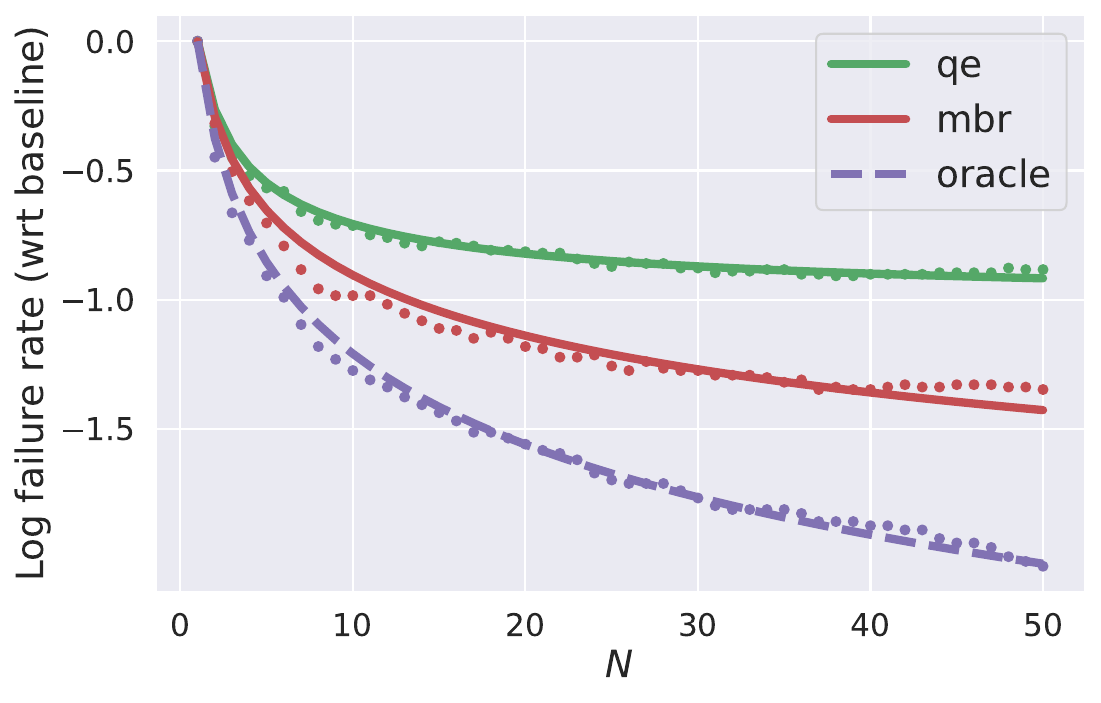}
\includegraphics[width=0.45\textwidth]{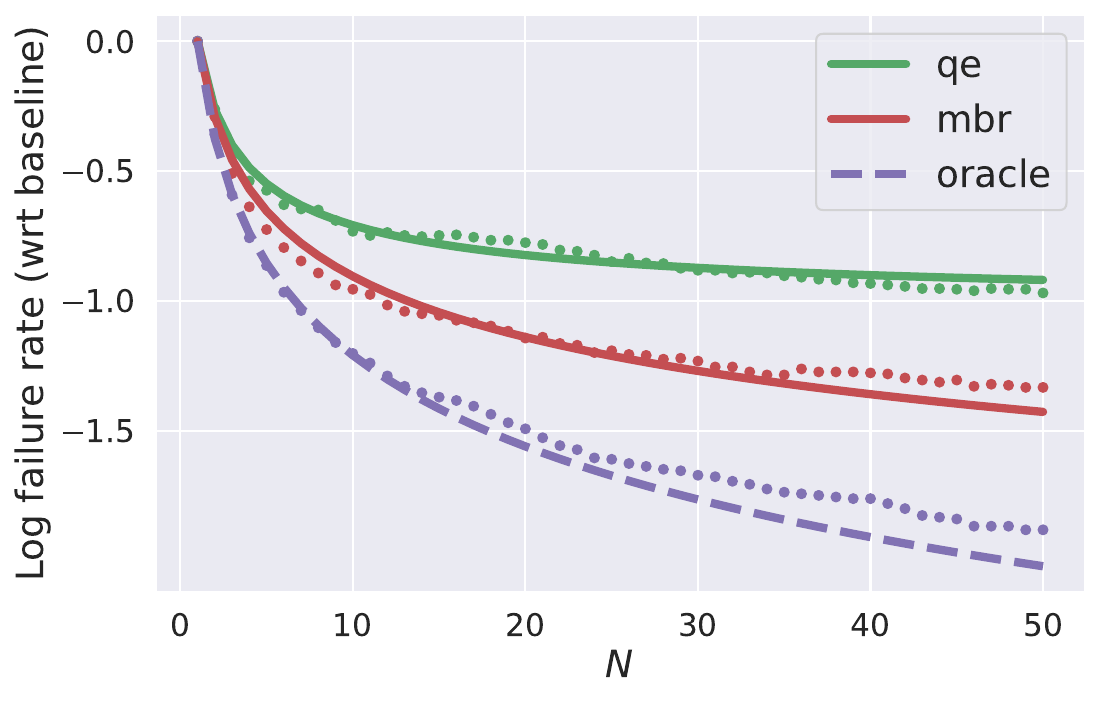}
\caption{
Log of the failure rate as a function of $N$. The empirical data is represented with dots (\textbf{left:} dev, \textbf{right:} test set) and our fitted models with solid and dashed lines (imperfect and perfect reranker, respectively).
In this case, we fit separate models for each language pair (\textbf{from top to bottom:} EN-PT, EN-ES, and EN-RU).}
\label{fig:mt-extra}
\end{figure}

\paragraph{Visualizations.}
In \cref{subsec:Machine translation} we obtained a single reranking law for the all language pairs; we now fit different models for each language pair.
\cref{fig:mt-extra} shows the log failure rate on the dev and test sets as a function of $N$ for EN-PT, EN-ES, and EN-RU. While the fits on the dev set are good, there is some degradation on the test set, especially for EN-ES (oracle and MBR decoding), possibly due to a shift in the distribution of scores/errors. We leave the investigation of more robust techniques and how to adapt to these cases for future work.

\subsection{Mathematical and commonsense reasoning}
\label{app:additional-experiments}

Our approach is fully general and can be useful in other domains other than code and language generation. In this subsection, we present additional experiments on mathematical and commonsense reasoning benchmarks, as prior work has shown that generating multiple hypotheses as an intermediate step is also advantageous in these scenarios \citep{wang2023selfconsistency}.

We use samples generated by \citet{aggarwal-etal-2023-lets} with code-davinci-002, a GPT-3-based model with 175 billion parameters \citep{brown2020language} which is part of the Codex series \citep{chen2021evaluating} (please refer to their Section~4 for more details; these samples were made publicly available by the authors at \href{https://github.com/Pranjal2041/AdaptiveConsistency}{https://github.com/Pranjal2041/AdaptiveConsistency}).
We apply self-consistency over diverse reasoning paths \citep{wang2023selfconsistency}, selecting the most frequent answer in the candidate set, and report results on the SVAMP \citep{patel-etal-2021-nlp} and StrategyQA \citep{geva2021did} datasets. Following \cref{subsec:Code generation}, we split the datasets in two equally sized parts to get development and test splits.

\begin{figure}[h]
\centering
\includegraphics[width=0.49\columnwidth]{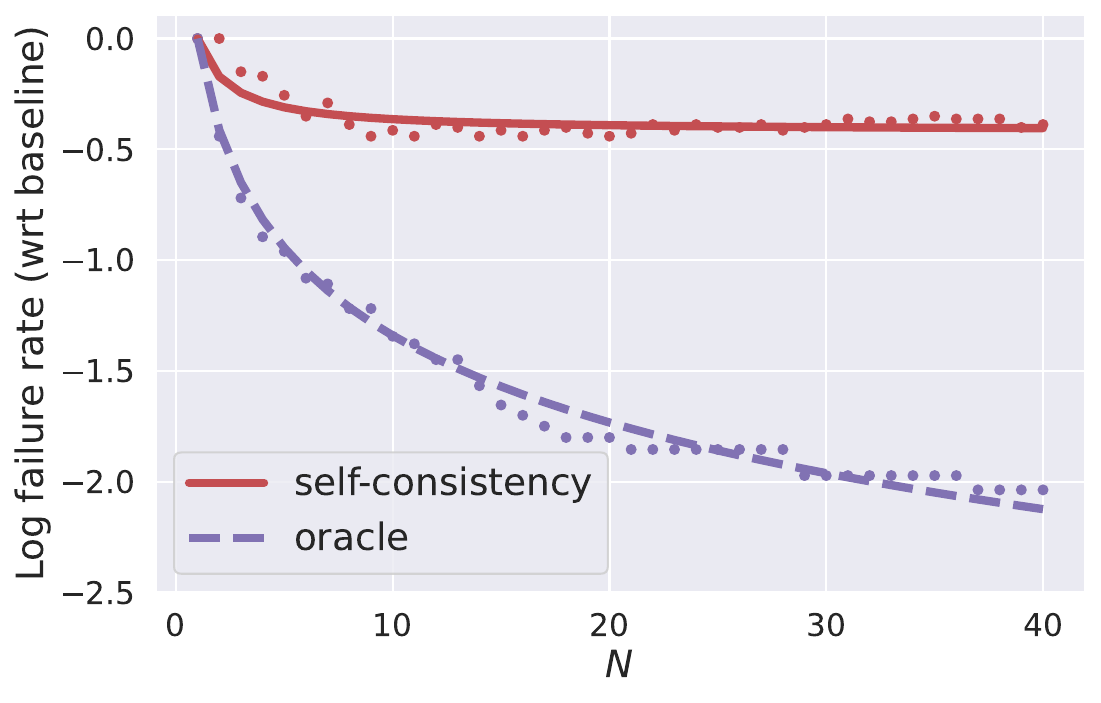}
\includegraphics[width=0.49\columnwidth]{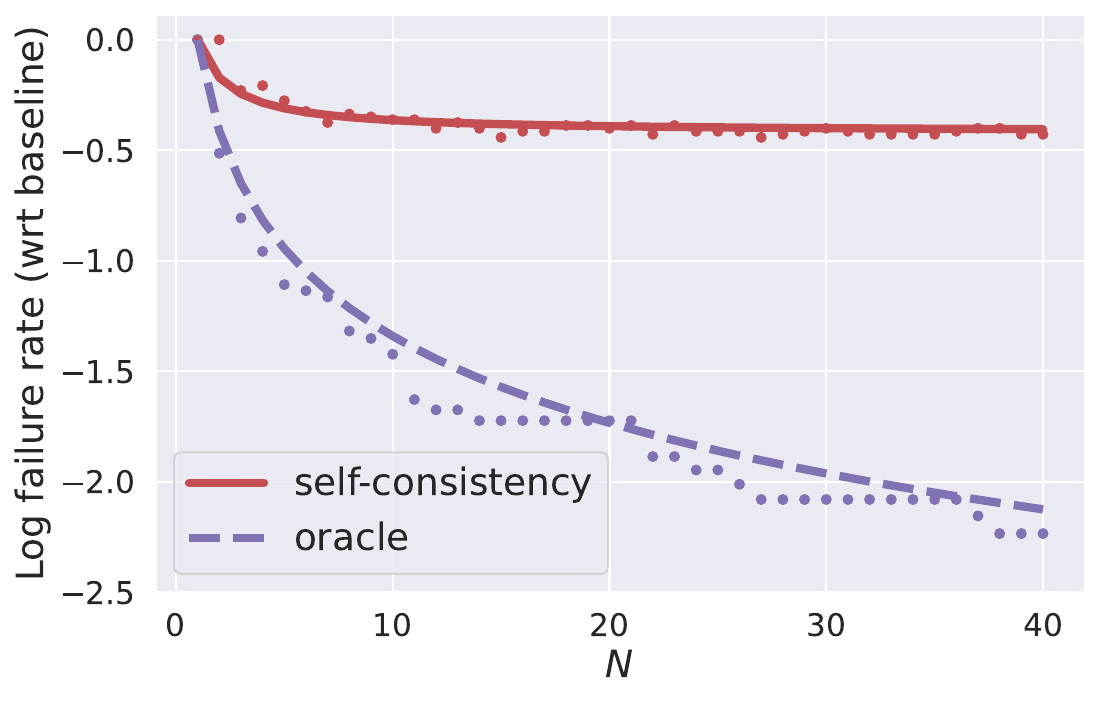}
\includegraphics[width=0.49\columnwidth]{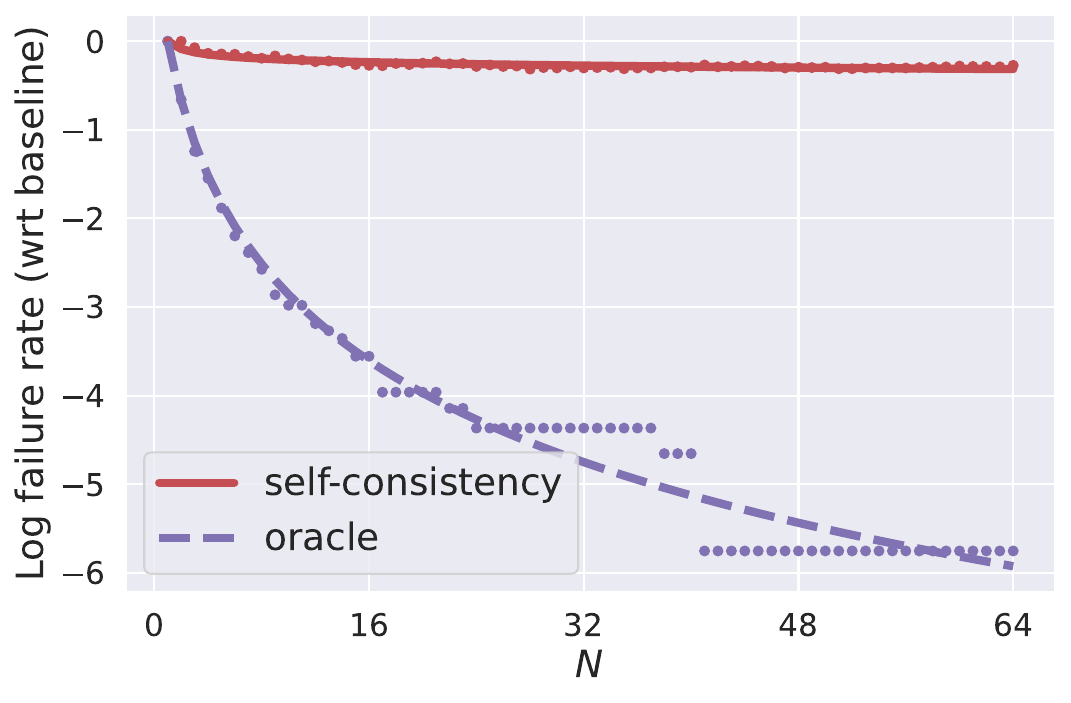}
\includegraphics[width=0.49\columnwidth]{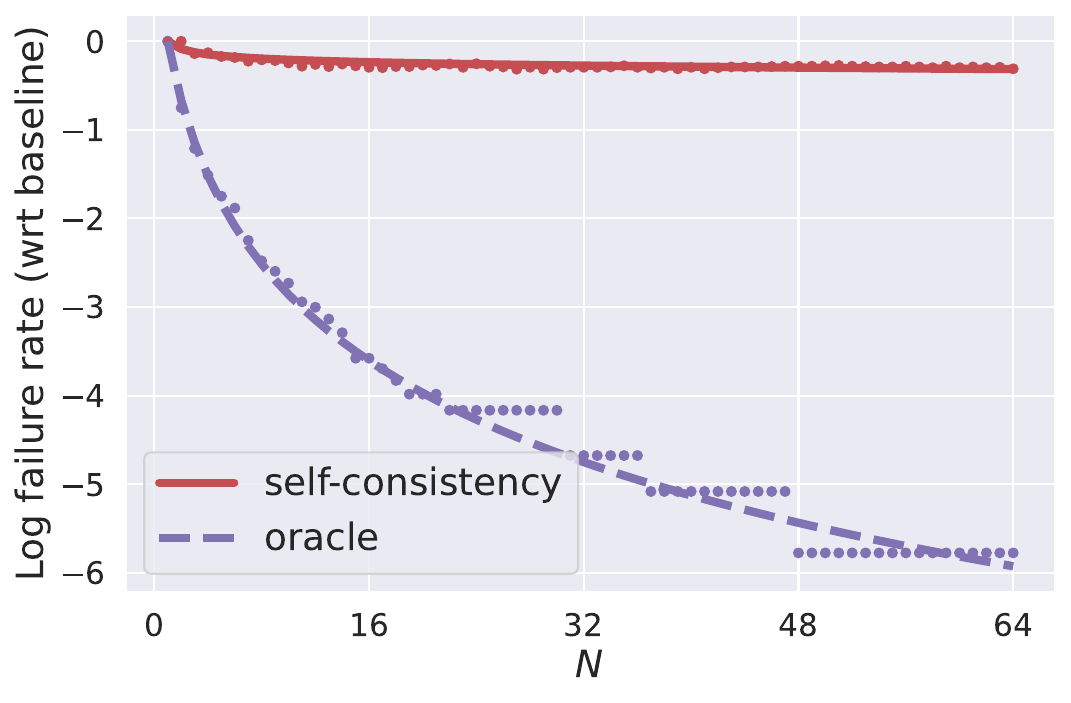}
\caption{Log of the failure rate as a function of $N$. The empirical data is represented with dots (\textbf{left:} dev, \textbf{right:} test set) and our fitted models with solid and dashed lines (imperfect and perfect reranker, respectively).
\textbf{Top:} mathematical reasoning on SVAMP . \textbf{Bottom:} commonsense reasoning on StrategyQA.}
\label{fig:additional-exps}
\end{figure}

Similarly to \cref{fig:exps}, \cref{fig:additional-exps} shows the log failure rate on the dev and test sets (left and right, respectively) as a function of $N$, confirming that the same trends hold also for these two additional tasks.

\subsection{Computing infrastructure}
\label{app:computing-infrastructure}

Our insfrastructure consists of $2$ machines, each equipped with $8$ NVIDIA RTX A6000 GPUs (46GB)  and $12$ Intel Xeon Gold 6348 CPUs (2.60GHz, 1TB RAM). The machines were used interchangeably, and all experiments were executed on a single GPU.


\end{document}